\documentclass{article}

\usepackage{microtype}
\usepackage{graphicx}
\usepackage{subfigure}
\usepackage{booktabs} 
\usepackage{multicol}
\usepackage{caption}

\usepackage{arydshln}

\usepackage{orcidlink}

\usepackage{xspace}
\newcommand{\ie}{i.\,e.,\xspace}
\newcommand{\eg}{e.\,g.,\xspace}

\usepackage{hyperref}

\usepackage[accepted]{icml2025}

\usepackage{amsmath}
\usepackage{amssymb}
\usepackage{mathtools}
\usepackage{amsthm}

\usepackage{cases}

\icmltitlerunning{iN2V: Bringing Transductive Node Embeddings to Inductive Graphs}

\begin{document}

\twocolumn[
\icmltitle{iN2V: Bringing Transductive Node Embeddings to Inductive Graphs}


\icmlsetsymbol{equal}{*}

\begin{icmlauthorlist}
\icmlauthor{Nicolas Lell\orcidlink{0000-0002-6079-6480}}{1}
\icmlauthor{Ansgar Scherp\orcidlink{0000-0002-2653-9245}}{1}
\end{icmlauthorlist}

\icmlaffiliation{1}{Research Group on Data Science and Big Data Analytics, Ulm University, Ulm, Germany}

\icmlcorrespondingauthor{Nicolas Lell}{Nicolas.Lell@uni-ulm.de}

\icmlkeywords{GNN, Graph, Inductive, Embedding, Node2Vec}

\vskip 0.3in
]



\printAffiliationsAndNotice{}  

\begin{abstract}

Shallow node embeddings like node2vec (N2V) can be used for nodes without features or to supplement existing features with structure-based information.
Embedding methods like N2V are limited in their application on new nodes, which restricts them to the transductive setting where the entire graph, including the test nodes, is available during training.
We propose inductive node2vec (iN2V), which combines a post-hoc procedure to compute embeddings for nodes unseen during training and modifications to the original N2V training procedure to prepare the embeddings for this post-hoc procedure.
We conduct experiments on several benchmark datasets and demonstrate that iN2V is an effective approach to bringing transductive embeddings to an inductive setting.
Using iN2V embeddings improves node classification by 1 point on average, with up to 6 points of improvement depending on the dataset and the number of unseen nodes.
Our iN2V is a plug-in approach to create new or enrich existing embeddings. 
It can also be combined with other embedding methods, making it a versatile approach for inductive node representation learning. 
Code to reproduce the results is available at \url{https://github.com/Foisunt/iN2V}.

\end{abstract}

\section{Introduction}

A graph neural network (GNN) may be trained without prior knowledge about the data it will encounter after deployment.
This is because, in real-world graphs, new nodes and edges appear or disappear over time.
For example, papers appear in citation networks, products are added to or removed from co-purchase graphs, and users join social platforms, creating new connections.
These scenarios align with the \textit{inductive setting} in graph learning, where test data is entirely unseen during training.
In contrast, the \textit{transductive setting} allows access to the entire graph during training, while the test node labels remain hidden.
From a broader perspective, the inductive setting resembles a single time step in a temporal graph learning task.
In temporal GNNs, the goal is to learn representations over multiple snapshots of a graph or a graph with temporal information attached to the nodes and edges, respectively~\cite{TGDK, tempgnns}. 

Various methods exist to compute node embeddings solely from graph edges~\cite{deepwalk, n2v, graphwave}.
This is important when node features are unavailable or for enriching existing features. 
When using text-based embeddings such as Bag of Words, some embeddings might be missing.
For example, due to out-of-vocabulary words, the Citeseer dataset~\cite{CoraCiteseer} contains 15 nodes with empty embeddings.  
Other reasons for new nodes include newly created social media accounts that yet lack user-provided information.

Absent node features pose a challenge for GNNs.
Message-passing GNNs such as GraphSAGE~\cite{sage} can infer missing information from neighboring nodes. 
MLP-based GNNs like Graph-MLP~\cite{graph-mlp} and GLNN~\cite{glnn} rely solely on node features, i.\,e., do not have message-passing, and cannot handle such cases.  
A common approach to addressing missing features is Feature Propagation~\cite{featureprop}.  
It propagates node features along graph edges, filling in missing features while preserving existing features.

We expand on this idea and introduce iN2V, a general and simple post-hoc approach to using trained embeddings to induce embeddings for nodes appearing in the inductive test set.
We modify the training process for the popular transductive embedding model node2vec~\cite{n2v} to foresee future embeddings.
We propose a simple but effective post-hoc procedure for propagating and updating embeddings for new nodes in the inductive setting, effectively enabling representation learning for unseen nodes.  
Unlike Feature Propagation (FP), our iN2V \textit{adapts} the embeddings of training nodes during propagation, a crucial feature to enable the inductive setting.
Since iN2V operates on N2V embeddings rather than raw node features provided by the datasets, it avoids relying on external node-specific information.  
Furthermore, it can be combined with any existing embeddings from the datasets, enriching the node representations and improving downstream task performance.
We evaluate the effectiveness of our iN2V embeddings on a range of homophilic and heterophilic datasets using the MLP and GraphSAGE models. 
Averaged over the other parameters, iN2V outperforms Feature Propagation by $1$ point on homophilic and $0.7$ points on heterophilic datasets, $1.3$ points when using MLP and $0.6$ points when using GraphSAGE as the classification model.
When using only the extended N2V embeddings, iN2V outperforms FP by $1.3$ points and by $0.6$ points when using both the extended N2V embeddings and graph features.
Finally, when using at most $20\%$ of the nodes for embedding generation and training, iN2V outperforms FP by $1.2$ points vs $0.8$ points when using at least $60\%$ of the nodes for training.
In summary, our contributions are:
\begin{itemize}
    \item Introduce iN2V, a simple and effective post-hoc method for extending trained node embeddings to unseen nodes.
    \item Enhance node2vec training with modifications that prepare their adaptability to inductive settings.
    \item Demonstrate performance gains, showing that both the inductive extension and modified training improve classification accuracy.
    \item Validate iN2V’s robustness, showing it remains effective even when trained on only $10\%$ of nodes; 
    in some cases outperforming using the original dataset features.
\end{itemize}

\section{Related Work}
\label{sec:relatedwork}

\subsection{Node Embeddings without Features}

Many node feature generation methods are based on random walks.
DeepWalk~\cite{deepwalk} generates random walks from a graph, treats each random walk as a sentence, and trains word2vec~\cite{Word2vec} embeddings on those random walks.
Building on DeepWalk, node2vec~\cite{n2v} introduces a biased random walk generator to better balance between locality and exploration
by giving distinct probabilities to return to the previous node, go to a node connected to the previous node, or visit a node not connected to the previous node.
LINE~\cite{line} generates two sets of embeddings independently and concatenates them afterward.
The first embedding optimizes that neighbors are similar, and the second embedding that nodes with many connections have similar embeddings.
Another approach to node embeddings is subgraph2vec~\cite{subgraph2vec}, which first generates rooted subgraphs for all nodes and then learns skip-gram embedding where the subgraphs of neighboring nodes are used as context for the current node.
While these approaches build on the idea that neighboring nodes should have more similar embeddings than distant nodes, struc2vec \cite{struc2vec} focuses on neighborhood degree patterns.
Random walks are done based on edge weights of a fully connected graph, with edge weights calculated from the similarity of the degree distribution of the neighborhood of each node.
GraphWave~\cite{graphwave} treats spectral graph wavelets as distributions to provide nodes with similar structural roles similar embeddings.
Sub2Vec~\cite{sub2vec} trains subgraph-level embeddings by applying paragraph2vec~\cite{paragraph2v} with additional random walks to better preserve the neighborhood and structural properties of the subgraphs.
There are also approaches for graph-level embeddings like graph2vec~\cite{graph2vec}.
It treats subgraphs as vocabulary and applies the doc2vec skip-gram training process.
Learning graph-level representations is less related to our work, which aims to learn features for nodes unseen during training.

RDF2vec~\cite{RDF2vec} is similar to DeepWalk but applied to Resource Description Framework (RDF) graphs.
It first converts an RDF graph into sequences and then trains word2vec~\cite{Word2vec} on them.
There are different follow-up works for RDF2vec; for example, \cite{hahn2024rdf2vec} used RDF2vec in a continual setup by sampling new walks starting from the new edges or entities.
Other embedding models used for Knowledge Graphs (KGs) aim to learn not only node but also edge or relation embeddings.
TransE~\cite{transE} embedded nodes and relations such that if there is a relation between two entities.
The first entity's embedding added with the relation embedding is trained to be close to the second entity's embedding.
Follow-up work~\cite{transH, complEx, rotatE} replaced the addition of real-valued vectors with other operations and other vector spaces, such as Hadamard product in a complex-valued vector space, and investigated regularizer and the effect of inverse relations on Knowledge Graph Completion (KGC) performance~\cite{canontensordecom}.
ReFactorGNNs~\cite{chen2022refactor} try to combine the good KGC performance of these factorization-based models with the ease of feature integration and inductive applicability of GNNs into a single model for KGC.
FedE~\cite{FedE} is a federated knowledge embedding framework that can use any knowledge graph embedding with multiple clients, each only having access to a part of the knowledge graph.
This is not applicable to an inductive setting, as the test graph is not seen at any time during the embeddings.

The Unifying Model~\cite{unifyingmodel} fits a Markov Random Field to a graph, which can be similar to label propagation or a linear GCN depending on the attributes used.
The model allows the sampling of new graphs from the training distribution, which differs from our task of providing embeddings to unseen nodes in graphs with no attributes.

\subsection{GNNs}
The most well-known graph neural network is GCN~\cite{gcn}, aggregating neighbors with weights based on their degree.
GraphSAGE~\cite{sage} modified GCN by considering the embedding of the current node separately from the neighbor aggregation and introducing sampling schemes to deal with large graphs. 
Other modifications of GCN used attention to assign different weights to neighbors~\cite{gat} or make deep models easier to train by adding different kinds of skip connections~\cite{gcnii, jk, nasc}.
For efficient models, besides reducing the number of message-passing layers~\cite{sgc}, some works trained MLPs without using the edges for inference.
Graph-MLP~\cite{graph-mlp} incorporated edge information by pulling neighboring embeddings closer together with a contrastive loss.
GLNN~\cite{glnn} and NOSMOG~\cite{nosmog} distill GNNs into MLPs.
NOSMOG additionally increased robustness to noise with adversarial feature augmentation and concatenates DeepWalk embeddings to the input for capturing more structured information.

Homophily is the characteristic of a graph in which neighboring nodes share the same class.
Heterophilic graphs, \ie graphs where neighbors usually belong to different classes, are an active area of research with work on how to measure homophily~\cite{linkx,adj_hom,unbiased} and heterophilic datasets~\cite{heter}.
Models that can better deal with heterophilic data use novel aggregations, consider multi-hop neighborhoods, distinguish homo- and heterophilic edges, or make the graph more homophilic using rewiring~\cite{h2gcn, linkx, esmlp, mixhop, hm, prgnn, acm, hete_rew}.

\section{Inductive N2V}
\label{sec:methods}

The principal idea of our inductive N2V (iN2V) algorithm is to simply assign each test node the average embedding of its neighbors from the training set.
This is repeated for multiple iterations to also deal with test nodes with longer distances to training nodes.
The N2V embedding training is modified so that the embedding of training nodes are optimally prepared to induce embeddings to new nodes only seen during testing.

\subsection{Notation and Formalization}
\label{sec:problemstatement}

Given a graph $G = (E, V)$ with node set $V \subset \mathbb{N}$ and edge set  $E \subseteq V \times V$, with disjoint training, validation, and test sets $V_{train}, V_{val}, V_{test} \subset V$.
$N(v) = \{x\in V \mid (v,x) \in E\}$ is the set of neighbors of $v$.
For each node $v \in V$, we want to train an embedding $h_v \in \mathbb{R}^d$. 
In the inductive setting, only the subgraph induced by $V_{train}$ is available for training the embeddings.
These embeddings then need to be extended to the remaining nodes $v_i \in V \setminus V_{train}$ in the validation and test set.
As we use existing benchmark datasets, our nodes also have classification labels $Y \in \mathbb{N}^{|V|}$ and existing node features $X \in \mathbb{R}^{|V|\times \hat d}$.

\begin{figure}
    \centering
    \includegraphics[width=0.85\linewidth]{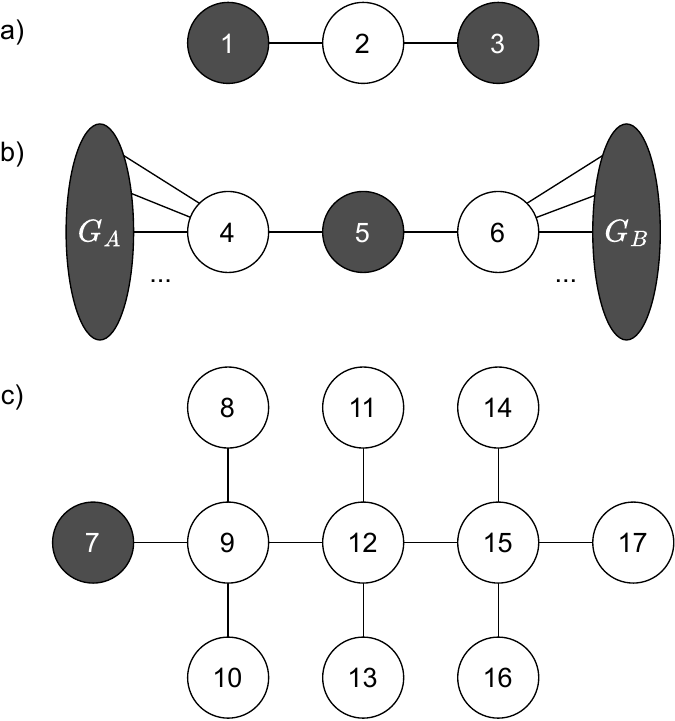}
    \caption{Three example graphs illustrate the post-hoc extensions to the white test nodes after obtaining embeddings for the gray training nodes in the inductive setup.
    Figure a) shows a simple example where node 2 obtains the average embedding of nodes 1 and 3.
    In Figure b) node 5 got a distant embedding during embedding training, but during the post-hoc extension it should be updated to move between the embeddings of the nodes from graphs $G_A$ and $G_B$.
    Finally, Figure c) illustrates how iN2V needs only four iterations to provide embeddings to all nodes.
    }
    \label{fig:example_graphs}
\end{figure}

\subsection{Example}

We motivate the different components of our post-hoc extension and illustrate their effect with the three example graphs a), b), and c) in Figure~\ref{fig:example_graphs}.
Training nodes are shown in gray, and test nodes in white.
The latter are hidden during training in the inductive case.
In graph a) it is quite straightforward that $v_2$ should obtain the average of the embeddings of $v_1$ and $v_3$.
In graph b), $G_A$ and $G_B$ are connected subgraphs of multiple training nodes with similar embeddings.
When following the averaging idea, $v_4$ gets an embedding that is close to the average embedding in $G_A$ but skewed towards the embedding of $v_5$.
Considering that the training embeddings were generated by N2V, the connected nodes in $G_A$ and $G_B$ got meaningful embeddings during training, while $v_5$ has a distant embedding as it has no neighbors in the training set and therefor only appeared as a negative sample during training.
Contrary to Feature Propagation~\cite{featureprop}, in this case, it is useful to allow the adaption of input (training) embeddings.
Following this line of thought while also maintaining some stability for the nodes with existing embeddings, each embedding should be a combination of itself and the average neighbor embedding.
When doing multiple iterations of such an averaging procedure, the embedding of $v_5$ moves in between the embeddings of nodes of $G_A$ and $G_B$.
However, too many iterations pose the challenge that all embeddings of individual nodes will converge to the average node embedding of that graph.
Graph c) illustrates the challenge of extending embeddings into longer sequences of test nodes.
When just averaging the neighborhood embedding for four iterations, the embedding of $v_{17} = v_7/4^3$ is close to zero.
Feature Propagation handles this by keeping the input embeddings fixed and iterating many times until convergence.
We already established the usefulness of adapting input embeddings in the example graph~b).
Therefore, we handle long sequences and high-degree nodes by considering only nodes that already have an embedding for averaging each iteration.
That means that after the first iteration $v_9 = v_7$, after the second iteration $v_8, v_{13}, v_{10}, v_9 = v_7$, and so on.
This leads to $v_{17} = v_7$ after only four iterations.

\subsection{Generating Inductive Embeddings}
We propose an iterative algorithm to extend trained embeddings to the unseen nodes.
Let $h^{(t)}_v$ be the embedding of node $v$ after $t$ iterations of our algorithm.
We use a lookup vector $s \in \{0,1\}^{|V|}$ with $s_v$ being the $v$-th element in the vector $s$ to keep track of which nodes already have embeddings and use $N_s(v) = \{ x\in N(v)\mid s_x = 1 \}$ to denote the set of neighbors which have an embedding.
The mean embedding of a set of nodes $S$ is $m^{(t)}_S = \frac{1}{|S|} \sum_{v\in S} h^{(t)}_v$.
For initialization, $h^{(0)}_{v}$ is set to the N2V embedding $h_u$ for training nodes $u \in V_{train}$ and to $0$ for nodes $w \in V \setminus V_{train}$ not from the training set.
The lookup vector is initialized with $s_u = 1$ and $s_w=0$.
Then $h^{t}$ is calculated from $h^{(t-1)}$ by:

\begin{subnumcases}{h^{(t)}_v = }
h^{(t-1)}_v & if $N_s(v) = \emptyset$ \label{1a} \\
\lambda h^{(t-1)}_v +  (1-\lambda) m^{(t-1)}_{N_s(v)} & if $s_v = 1$  \label{1b} \\ 
m^{(t-1)}_{N_s(v)} & else \label{1c}
\end{subnumcases}

This means that if $v$ has no neighbor with an embedding, $h_v$ does not change (\ref{1a}).
If both $v$ and at least one neighbor of $v$ have an embedding, we calculate the convex combination of $h_v$ and the mean neighbor embedding $m_{N_s(v)}$ (\ref{1b}).
Note, for $\lambda = 1$, the embedding of a node will not change once it is set, and with $\lambda \le 1$ all embeddings will be updated depending on their respective neighborhoods.
If $v$ does not have an embedding but at least one neighbor has an embedding, we set $v$'s embedding to the mean neighbor embedding (\ref{1c}).
This is done for multiple iterations.
After each iteration, $s$ is updated by setting entries for nodes that got an embedding to $1$.
We do enough iterations such that each node with a path to at least one training node gets an embedding and additional $delay$-many iterations to update the embeddings of nodes like $v_5$ in the example Figure~\ref{fig:example_graphs}~b).

\subsection{Boosting Inductive Performance}

We propose two different approaches which modify the training to improve the generation of inductive embeddings.

\paragraph{Sampling-based}
To promote embeddings that are better suited to the inductive extension of embeddings, we simulate a simple version of the post-hoc extensions during training.
In each epoch, some features are replaced by their mean neighborhood embedding with probability $r$.

\begin{equation*}
h_v =
\begin{cases}
m_{N(v)} & \text{with probability } r\\
h_v & \text{else}    
\end{cases}
\end{equation*}

\paragraph{Loss-based}
In addition to the sampling-based approach, we also introduce a loss-based approach to prepare the embeddings for our inductive extension.
When extending the trained embeddings to the inductive nodes during inference, we set the embedding of new nodes to their mean neighborhood embedding.
The first loss promotes this relationship in the trained embeddings by pulling a node's own embedding closer to its mean neighborhood embedding:

\begin{equation*}
\mathcal{L}_{close}(v) = -\log( \sigma(h_v \cdot m_{N(v)})) \,.
\end{equation*}

A trivial solution to minimize this loss would be to assign identical embeddings to all neighboring nodes, so we add a second loss which promotes diversity in embeddings of the individual neighbors of each node:

\begin{equation*}
\mathcal{L}_{div}(v) = \frac{1}{|N(v)|^2} \sum_{u,w \in N(v)} \operatorname{sim}(h_u, h_w) \,,
\end{equation*}

where $\operatorname{sim}$ is cosine similarity.
The final loss for iN2V is

\begin{equation*}
\mathcal{L}(v) = \mathcal{L}_{n2v}(v) + \alpha \cdot \mathcal{L}_{close}(v) + \beta \cdot \mathcal{L}_{div}(v) 
\end{equation*}

with hyperparameters $\alpha$ and $\beta$.
The N2V loss $\mathcal{L}_{n2v}$ is calculated using random walks.
In each epoch, every node appears on average $ \text{walks\ per\ node}$ times $\text{walk\ length}$ often in these random walks.
The random walks are batched for training, but when the same node appears multiple times in a batch, the additional calculations for $\mathcal{L}_{close}$ and $\mathcal{L}_{div}$ are redundant.
To reduce these redundant calculations, we sample the nodes to calculate $\mathcal{L}_{close}$ and $\mathcal{L}_{div}$ independent from the random walks such that each node is the center node for these losses once per epoch.

\section{Experimental Apparatus}
\label{sec:experimentalapparatus}

\subsection{Datasets}
\label{sec:datasets}

We use the Cora~\cite{CoraCiteseer}, CiteSeer~\cite{CoraCiteseer}, and PubMed~\cite{pubmed} citation graphs, the Computers~\cite{shchur} and Photo~\cite{shchur} co-purchase graphs, and the WikiCS Wikipedia page graph~\cite{wikics}.
These graphs are homophilic, \ie neighboring nodes usually share the same class.
The following graphs are more heterophilic, \ie neighboring nodes usually do not share the same class.
Actor~\cite{geomgcn} is a Wikipedia co-occurrence graph, Amazon-ratings~\cite{heter} is a co-purchase graph, and Roman-empire~\cite{heter} is a text-based graph.
Table~\ref{tab:dataset} shows more details about the datasets.
We report adjusted homophily as it accounts for class imbalance~\cite{adj_hom}.

\begin{table}
\centering    
\caption{Graph features, nodes, edges, classes, and adjusted homophily. Upper: homophilic, bottom: heterophilic graphs. }
\label{tab:dataset}
\resizebox{0.99\linewidth}{!}{
\begin{tabular}{lrrrrrr} \toprule
Dataset     & $|X|$    & $|V|$     &  $|E|$     & $|C|$ & $hom_{adj}$ \\ \midrule
Cora        & $1\,433$ &  $2\,708$ &  $10\,556$ &   $7$ & $0.77$ \\
CiteSeer    & $3\,703$ &  $3\,327$ &   $9\,104$ &   $6$ & $0.67$ \\
PubMed      &    $500$ & $19\,717$ &  $88\,648$ &   $3$ & $0.69$ \\
Computers   &    $767$ & $13\,752$ & $491\,722$ &  $10$ & $0.68$ \\
Photo       &    $745$ &  $7\,650$ & $238\,162$ &   $8$ & $0.79$ \\ 
WikiCS      &    $300$ & $11\,701$ & $431\,726$ &  $10$ & $0.58$\\ 
\hdashline
Actor       &    $932$ &  $7\,600$ &  $30\,019$ &   $5$ & $0.01$ \\ 
Amazon-R.   &    $300$ & $24\,492$ &  $93\,050$ &   $5$ & $0.14$ \\
Roman-E.    &    $300$ & $22\,662$ &  $32\,927$ &  $18$ & $-0.05$ \\ 
\bottomrule
\end{tabular} }
\end{table}

\subsection{Procedure}
\label{sec:procedure}

We investigate how well iN2V works for differently sized data splits into training and unseen validation or test nodes.
For all datasets, we use 5 splits of different sizes that always utilize the full dataset and have a validation and test set of the same size.
The training set sizes are $10\%$, $20\%$, $40\%$, $60\%$, and $80\%$, with respective validation and test set sizes of $45\%$, $40\%$, $30\%$, $20\%$, and $10\%$ of all nodes.

First, we prepare $10$ splits from different random seeds for each of the five split sizes.
Then we train N2V inductively on the training subgraphs.
For the N2V hyperparameter search, we use a grid search on three of the ten splits.
The embeddings from the training set are extended to the validation set using iN2V.
The final N2V hyperparameters are chosen based on the validation accuracy of logistic regression on these embeddings.
These hyperparameters are used to train and store embeddings for all $10$ splits.
Then we evaluate the embeddings using MLP and GraphSAGE~\cite{sage}.
Additionally, we investigate the concatenation of the extended N2V embeddings with the original graph features.

Regarding the extended embeddings, we compare different setups of iN2V.
The first is the ``frozen'' setup ($\lambda =1$), where embeddings do not change after being set.
The second one is the best post-hoc setup, where $\lambda$ and $delay$ are searched as hyperparameters.
The third and fourth ones combine the sampling-based and loss-based modifications to N2V with the best post-hoc setup.
We compare these results with different baselines.
The comparable baselines that use the same information as iN2V are plain N2V used inductively, \ie only training nodes have embeddings, and using Feature Propagation~\cite{featureprop} to extend N2V embeddings to the test nodes.
Using the original graph features and training N2V embeddings in a transductive setup are two more baselines.
They are not directly comparable with iN2V as they use more information but are nevertheless useful for perspective.
Additional experiments with other GNNs can be found in Appendix~\ref{appx:gat_gin}.

\subsection{Hyperparameter Optimization}
\label{sec:hyperparameteroptimization}

We make all datasets undirected.
For N2V, we use a context size of $10$ for positive samples and $1$ negative per positive sample.
We use a batch size of $128$ and early stopping with patience of $50$ epochs.
For every epoch, we sample 10 walks of length $20$ per node.
We do grid search over all combinations of $p$ and $q \in \{ 0.2,\, 1,\, 5\}$, embedding size $d \in \{64,\, 256\}$, and learning rate $\in \{0.1,\,  0.01,\,  0.001\}$.
For the sampling based method, we try $r \in \{0.2,\,  0.4,\,  0.6,\,  0.8\}$.
The loss weights $\alpha \in \{0,\,  0.1,\,  1,\,  10\}$ and $\beta \in \{0,\, 0.001,\, 0.01,\, 0.1\}$ are tuned separately from $r$. 
For Feature Propagation, we search the number of iterations in $\{10,\,  20,\,  40,\,  60\}$.

For MLP and GraphSAGE, we use grid search for the full $10$ seeds per split.
We search over all combination of number of layer $\in \{1,\,  ...\, 5\}$, hidden size $\in \{64,\,  512\}$, learning rate $\in \{0.01,\,  0.001\}$, weight decay $\in \{0,\,  0.0001,\,  0.01\}$, dropout $\in \{0.2,\,  0.5, \, 0.8\}$, and whether to use jumping knowledge~\cite{jk} connections.

\section{Results and Discussion}
\label{sec:results}

\subsection{Key Results}

\paragraph{Comparison of iN2V vs. Baselines}

Table \ref{tab:res_sage} shows the performance of GraphSAGE using iN2V embeddings and compares them to normal N2V in the inductive setting (test nodes have no features) and N2V embeddings extended by Feature Propagation.
We also report performance on the original graph features and transductive N2V embeddings to indicate how close the baselines and iN2V are to the ideal case of no missing nodes or features.
We can see that iN2V outperforms the comparable baselines in most cases.
The notable exceptions are the Actor and Roman-empire datasets, where N2V embeddings generally perform poorly.

In some cases, using the iN2V embeddings outperforms the original features, \eg in the $10\%$ training splits of Cora, Computers, and Amazon-Ratings.
This implies that neighborhood information is more important for these datasets than the external information from the original features and that the original features do not have sufficient neighborhood information.
Appendix~\ref{appendix:res_full} provides tables with the full results, including the $60\%$ train split for GraphSAGE and all MLP results.
We observe for some datasets that the iN2V embeddings outperform the original features for more splits when using MLP as a model, as it can not generate neighborhood information internally as GraphSAGE does.

\begin{table}[]
\centering
\caption{Accuracy of best iN2V variant vs baseline embeddings. The underlying model is GraphSAGE. Gray numbers are not directly comparable as they use additional information, \ie original features, or the transductive setup.}
\resizebox{\linewidth}{!}{
\begin{tabular}{lrrrr} \toprule
& \multicolumn{4}{l}{Percentage of training data} \\ \midrule
 Dataset & 10\% & 20\% & 40\% & 80\% \\ \midrule

\multicolumn{5}{l}{Cora} \\ \hline
N2V (inductive) & $42.18_{3.52}$ & $59.91_{4.15}$ & $75.07_{1.92}$ & $\mathbf{84.50}_{1.44}$ \\
Feature Propagation & $77.91_{2.62}$ & $79.48_{2.19}$ & $81.03_{1.85}$ & $84.13_{2.35}$ \\
\textbf{iN2V} (own) & $\mathbf{78.88}_{1.45}$ & $\mathbf{80.94}_{1.58}$ & $\mathbf{83.30}_{1.09}$ & $84.46_{2.08}$ \\ \hline
\textcolor{gray}{Original features} & \textcolor{gray}{$75.27_{2.63}$} & \textcolor{gray}{$83.37_{1.17}$} & \textcolor{gray}{$86.23_{1.77}$} & \textcolor{gray}{$87.05_{1.20}$} \\
\textcolor{gray}{N2V (transductive)} & \textcolor{gray}{$79.25_{1.45}$} & \textcolor{gray}{$81.66_{1.29}$} & \textcolor{gray}{$83.81_{0.95}$} & \textcolor{gray}{$86.01_{1.99}$} \\ \midrule

\multicolumn{5}{l}{Citeseer} \\ \hline
N2V (inductive) & $34.17_{3.43}$ & $42.76_{2.15}$ & $56.12_{3.24}$ & $68.89_{2.61}$ \\
Feature Propagation & $56.85_{1.84}$ & $60.53_{2.03}$ & $\mathbf{63.03}_{2.21}$ & $\mathbf{69.76}_{1.82}$ \\
\textbf{iN2V} (own) & $\mathbf{57.88}_{0.91}$ & $\mathbf{60.78}_{1.78}$ & $63.02_{1.88}$ & $68.92_{2.07}$ \\ \hline
\textcolor{gray}{Original features} & \textcolor{gray}{$69.85_{1.42}$} & \textcolor{gray}{$72.86_{0.98}$} & \textcolor{gray}{$74.93_{1.56}$} & \textcolor{gray}{$76.82_{1.77}$} \\
\textcolor{gray}{N2V (transductive)} & \textcolor{gray}{$57.14_{1.51}$} & \textcolor{gray}{$61.33_{1.09}$} & \textcolor{gray}{$66.45_{1.60}$} & \textcolor{gray}{$72.76_{2.54}$} \\ \midrule

\multicolumn{5}{l}{Pubmed} \\ \hline
N2V (inductive) & $66.02_{5.57}$ & $74.73_{2.36}$ & $79.84_{1.78}$ & $\mathbf{82.61}_{0.52}$ \\
Feature Propagation & $76.37_{0.62}$ & $77.72_{0.50}$ & $80.74_{0.72}$ & $82.43_{0.86}$ \\
\textbf{iN2V} (own) & $\mathbf{79.93}_{0.50}$ & $\mathbf{80.80}_{0.46}$ & $\mathbf{82.14}_{0.43}$ & $82.59_{0.63}$ \\ \hline
\textcolor{gray}{Original features} & \textcolor{gray}{$85.95_{0.47}$} & \textcolor{gray}{$86.99_{0.28}$} & \textcolor{gray}{$88.32_{0.48}$} & \textcolor{gray}{$89.85_{0.56}$} \\
\textcolor{gray}{N2V (transductive)} & \textcolor{gray}{$81.36_{0.49}$} & \textcolor{gray}{$82.20_{0.50}$} & \textcolor{gray}{$83.22_{0.34}$} & \textcolor{gray}{$83.66_{0.65}$} \\ \midrule

\multicolumn{5}{l}{Computers} \\ \hline
N2V (inductive) & $77.64_{2.81}$ & $84.44_{0.84}$ & $87.18_{0.77}$ & $89.35_{0.70}$ \\
Feature Propagation & $82.79_{0.64}$ & $86.43_{0.63}$ & $89.40_{0.45}$ & $90.87_{0.64}$ \\
\textbf{iN2V} (own) & $\mathbf{88.36}_{0.58}$ & $\mathbf{89.67}_{0.40}$ & $\mathbf{90.84}_{0.37}$ & $\mathbf{91.38}_{0.51}$ \\ \hline
\textcolor{gray}{Original features} & \textcolor{gray}{$87.52_{0.48}$} & \textcolor{gray}{$89.76_{0.40}$} & \textcolor{gray}{$91.12_{0.20}$} & \textcolor{gray}{$91.50_{0.48}$} \\
\textcolor{gray}{N2V (transductive)} & \textcolor{gray}{$89.18_{0.38}$} & \textcolor{gray}{$90.16_{0.44}$} & \textcolor{gray}{$90.77_{0.39}$} & \textcolor{gray}{$91.16_{0.56}$} \\ \midrule

\multicolumn{5}{l}{Photo} \\ \hline
N2V (inductive) & $85.48_{1.28}$ & $87.73_{1.40}$ & $90.98_{0.71}$ & $92.21_{0.96}$ \\
Feature Propagation & $87.43_{1.09}$ & $90.14_{0.34}$ & $91.57_{0.42}$ & $92.95_{0.79}$ \\
\textbf{iN2V} (own) & $\mathbf{90.51}_{0.72}$ & $\mathbf{91.70}_{0.44}$ & $\mathbf{92.37}_{0.46}$ & $\mathbf{93.08}_{0.77}$ \\ \hline
\textcolor{gray}{Original features} & \textcolor{gray}{$93.74_{0.42}$} & \textcolor{gray}{$94.59_{0.37}$} & \textcolor{gray}{$95.27_{0.38}$} & \textcolor{gray}{$95.59_{0.77}$} \\
\textcolor{gray}{N2V (transductive)} & \textcolor{gray}{$91.29_{0.41}$} & \textcolor{gray}{$92.29_{0.30}$} & \textcolor{gray}{$92.90_{0.45}$} & \textcolor{gray}{$93.33_{0.71}$} \\ \midrule

\multicolumn{5}{l}{WikiCS} \\ \hline
N2V (inductive) & $67.78_{2.68}$ & $74.22_{1.65}$ & $78.21_{0.69}$ & $81.62_{0.86}$ \\
Feature Propagation & $74.77_{2.27}$ & $78.04_{1.00}$ & $80.17_{0.68}$ & $81.93_{1.08}$ \\
\textbf{iN2V} (own) & $\mathbf{78.91}_{0.61}$ & $\mathbf{80.19}_{0.70}$ & $\mathbf{81.28}_{0.61}$ & $\mathbf{82.37}_{1.01}$ \\ \hline
\textcolor{gray}{Original features} & \textcolor{gray}{$80.75_{0.64}$} & \textcolor{gray}{$82.56_{0.81}$} & \textcolor{gray}{$84.28_{0.55}$} & \textcolor{gray}{$85.88_{0.70}$} \\
\textcolor{gray}{N2V (transductive)} & \textcolor{gray}{$79.75_{0.41}$} & \textcolor{gray}{$80.93_{0.64}$} & \textcolor{gray}{$81.88_{0.55}$} & \textcolor{gray}{$82.81_{0.68}$} \\ \midrule

\multicolumn{5}{l}{Actor} \\ \hline
N2V (inductive) & $25.14_{1.09}$ & $\mathbf{25.56}_{0.90}$ & $\mathbf{25.54}_{1.15}$ & $25.68_{1.42}$ \\
Feature Propagation & $\mathbf{25.22}_{1.29}$ & $25.39_{1.06}$ & $25.14_{0.59}$ & $25.05_{1.34}$ \\
\textbf{iN2V} (own) & $25.18_{0.97}$ & $25.50_{0.70}$ & $25.40_{0.99}$ & $\mathbf{25.79}_{2.27}$ \\ \hline
\textcolor{gray}{Original features} & \textcolor{gray}{$31.77_{0.71}$} & \textcolor{gray}{$33.84_{0.91}$} & \textcolor{gray}{$36.48_{0.54}$} & \textcolor{gray}{$36.71_{1.23}$} \\
\textcolor{gray}{N2V (transductive)} & \textcolor{gray}{$25.50_{0.76}$} & \textcolor{gray}{$25.56_{0.95}$} & \textcolor{gray}{$25.41_{0.74}$} & \textcolor{gray}{$24.55_{1.70}$} \\ \midrule

\multicolumn{5}{l}{Amazon-ratings} \\ \hline
N2V (inductive) & $37.47_{0.47}$ & $40.69_{0.63}$ & $44.71_{0.85}$ & $49.47_{1.07}$ \\
Feature Propagation & $38.97_{0.76}$ & $41.68_{0.76}$ & $\mathbf{45.72}_{0.75}$ & $50.04_{1.42}$ \\
\textbf{iN2V} (own) & $\mathbf{40.02}_{0.79}$ & $\mathbf{42.01}_{0.48}$ & $45.48_{0.57}$ & $\mathbf{50.38}_{2.22}$ \\ \hline
\textcolor{gray}{Original features} & \textcolor{gray}{$39.20_{1.06}$} & \textcolor{gray}{$41.66_{0.70}$} & \textcolor{gray}{$48.07_{0.58}$} & \textcolor{gray}{$57.34_{0.97}$} \\
\textcolor{gray}{N2V (transductive)} & \textcolor{gray}{$41.82_{0.66}$} & \textcolor{gray}{$43.69_{0.52}$} & \textcolor{gray}{$46.31_{0.78}$} & \textcolor{gray}{$49.80_{0.94}$} \\ \midrule

\multicolumn{5}{l}{Roman-empire} \\ \hline
N2V (inductive) & $\mathbf{13.96}_{0.35}$ & $\mathbf{14.10}_{0.32}$ & $\mathbf{15.77}_{0.76}$ & $16.81_{2.94}$ \\
Feature Propagation & $13.23_{1.44}$ & $13.89_{0.44}$ & $15.49_{0.67}$ & $\mathbf{21.97}_{1.25}$ \\
\textbf{iN2V} (own) & $13.79_{0.39}$ & $13.86_{0.29}$ & $14.48_{0.63}$ & $18.55_{1.23}$ \\ \hline
\textcolor{gray}{Original features} & \textcolor{gray}{$66.09_{0.75}$} & \textcolor{gray}{$70.28_{0.63}$} & \textcolor{gray}{$74.41_{0.47}$} & \textcolor{gray}{$82.90_{1.09}$} \\
\textcolor{gray}{N2V (transductive)} & \textcolor{gray}{$13.82_{0.22}$} & \textcolor{gray}{$13.86_{0.37}$} & \textcolor{gray}{$15.35_{1.53}$} & \textcolor{gray}{$27.46_{1.36}$} \\ \bottomrule 
\end{tabular}}
    \label{tab:res_sage}
\end{table}

\paragraph{Homophilic vs Heterophilic Datasets}

While we generally see quite good performance of models using N2V embeddings on homophilic datasets, this is not the case for heterophilic ones.
This is expected since N2V provides similar embeddings for neighbors, which matches the homophilic label structure.
On Actor, the models using N2V embeddings only learn to predict the largest class ($25.86\%$).
On Roman-empire, the models only predict the largest class ($13.96\%$) when using a few training nodes and become slightly better when using most of the data during training.
Roman-empire is the dataset with the lowest average degree in our comparison. This means the graph is prone to splitting into many disconnected subgraphs in the inductive setting.
This makes this dataset challenging for training any random-walk-based embeddings.
It also explains the rise in performance for the largest training split.
Amazon-ratings is an interesting exception to this trend because even though it has a low adjusted homophily of $0.14$, N2V embeddings seem to work as well as on the homophilic datasets and even outperform using the original features for small training splits.
Table~\ref{tab:full_mlp_base} paints a similar picture when using MLP instead of GraphSAGE, with the difference that even for $80\%$ training data, the MLP does not perform better than just predicting the largest class for the Roman-empire dataset.

\paragraph{Influence of GNN Choice and N2V Modifications}

Table~\ref{tab:emb_cat} shows the performance aggregated over all datasets and train splits.
The first two columns show the performance impact of the loss-based and sampling-based modifications to N2V when using only those embeddings as input for the GNNs.
Our post-hoc extension, which can adapt input embeddings, outperforms Feature Propagation and the frozen post-hoc variant.
This shows that adapting existing features is an important capability of iN2V.
The loss-based and sampling-based modifications to N2V training provide a small boost to performance.
When we compare the results for MLP vs GraphSAGE, we see that the differences are bigger when using an MLP.
Interestingly, MLP outperforms GraphSAGE when using post-hoc embeddings.
This is the case because GraphSAGE's main advantage over MLP is its ability to aggregate neighborhood information internally.
N2V already encodes this information in the embeddings, and additionally, our post-hoc extension is similar to the aggregation performed by models like GraphSAGE.
This effect might also be reinforced by the fact that we use logistic regression on the embeddings for the N2V hyperparameter selection, which is closer to an MLP than it is to GraphSAGE.
Detailed results per dataset and split are shown in Appendix~\ref{appendix:res_full}.

\begin{table}[]
    \centering
        \caption{Effect of model and input, test accuracy averaged over all datasets and splits.}
    \label{tab:emb_cat}
\begin{tabular}{lrrrr}
\toprule
Input & \multicolumn{2}{c}{embed} & \multicolumn{2}{c}{embed $||$ feat.} \\
Model & MLP & SAGE & MLP & SAGE \\
\midrule
N2V (inductive)  & $27.91$ & $59.45$ & $68.72$ & $73.51$ \\
Feature Propagation  & $62.97$ & $63.20$ & $73.17$ & $74.09$ \\
frozen ($\lambda=1$)  & $61.74$ & $62.88$ & $72.40$ & $73.46$ \\
post-hoc  & $64.38$ & $63.75$ & $73.99$ & $74.24$ \\
p-h w losses  & $64.57$ & $63.95$ & $74.04$ & $74.32$ \\
p-h w sampling & $64.42$ & $63.83$ & $73.95$ & $74.02$  \\ \hline
\textcolor{gray}{Original features} & \textcolor{gray}{$71.01$} & \textcolor{gray}{$74.73$} & \textcolor{gray}{$71.01$} & \textcolor{gray}{$74.73$} \\
\textcolor{gray}{N2V(transductive)} & \textcolor{gray}{$63.89$} & \textcolor{gray}{$65.07$} & \textcolor{gray}{$74.15$} & \textcolor{gray}{$75.57$} \\
\bottomrule
\end{tabular}
\end{table}

\paragraph{Combining N2V embeddings and Original Features}

For datasets that already have features, N2V embeddings can be used to provide additional structural information to GNNs.
The third and fourth columns of Table~\ref{tab:emb_cat} show our results in an aggregated way, while per dataset results can be found in Appendix~\ref{appx:cat}.
Using both the N2V embeddings and original features as input for the models increases overall performance by about $10$ points compared to just using N2V embeddings.
The increase is bigger for the N2V (inductive) baseline, as the test nodes do not have N2V embeddings in that scenario.
When just using N2V embeddings, MLP has a slightly higher average performance than GraphSAGE in most setups; this switches to GraphSAGE having a slight lead over MLP when using both as input.

Compared to the original features baseline, GraphSAGE actually loses an average of $.5$ performance points when also using the N2V as input, whereas MLP gains $3$ points.
This again shows that MLP can benefit from more structural information while GraphSAGE is already capable of aggregating this information by itself.
The increase heavily depends on the dataset; see detailed per-dataset results in Appendix~\ref{appx:cat}.
Both models gain over $50$ points on Roman-empire and have only smaller gains on the other datasets.
These dataset-specific differences are explained by the usefulness of N2V embeddings vs the usefulness of the original features.
On Roman-empire, the graph is close to a sequence of words with few additional edges; the neighborhood information encoded by N2V embeddings does not bring much useful information.

\paragraph{Influence of Amount of Training Data}

\begin{figure}
    \centering
    \includegraphics[width=\linewidth]{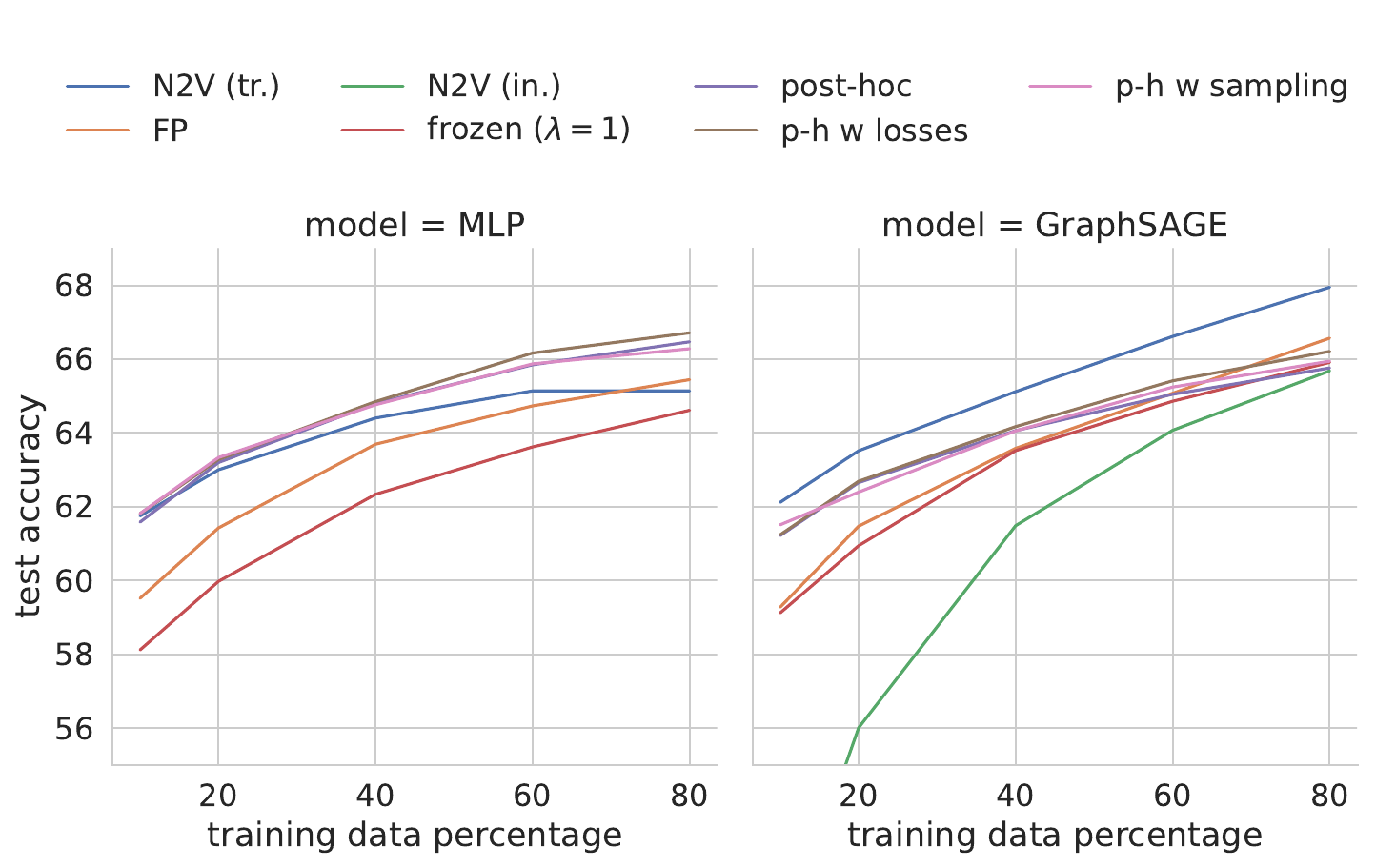}
    \caption{Influence of amount of training data. Results are averaged over all datasets.}
    \label{fig:split_effect}
\end{figure}

Figure~\ref{fig:split_effect} visualizes the effect of the training set size averaged over all datasets.
In general, performance increases with more training data.
The only exception is N2V in the inductive setting when using MLP, as the test nodes do not have embeddings, and performance stays at a random guessing level.
GraphSAGE can compensate for the missing test embeddings. 
While starting with low performance for little training data, it is close to the methods that actually extend embeddings to test nodes for $80\%$ training data.
The MLP model with Feature Propagation or frozen post-hoc ($\lambda=1$) consistently performs below the other post-hoc variants. 
GraphSAGE with Feature Propagation and frozen post-hoc catch up and outperform some other variants when using $80\%$ training data.
Another interesting observation is that for MLP, the transductive N2V starts in line with the post-hoc variants but rises less when the training data increases.
For GraphSAGE, transductive N2V already starts higher than the post-hoc variants and even widens the lead when the amount of training data increases.

\paragraph{Applying of Loss and Sampling Modification with Feature Propagation}

While our modifications to the N2V training procedure are motivated by our post-hoc extension, these two parts act independently. 
The loss-based and sampling-based modifications change the generated N2V embeddings, and the post-hoc algorithm extends these independent of their exact values.
This means that instead of using our post-hoc algorithm, we can also use Feature Propagation to extend the modified N2V embeddings.
Table~\ref{tab:fp+} shows the results of this experiment averaged over all datasets.
Feature Propagation gains around $.05$ points with the loss modification of the embeddings for both models.
The sampling-based modification does not change the MLP performance, but it reduces the GraphSAGE performance by $.2$ points.
This is lower than the average of $.2$ points gained with post-hoc with the losses and $.06$ gained by the sampling-based modification in Table~\ref{tab:emb_cat} and shows the synergy of our post-hoc methods with the N2V modifications.

\begin{table}[]
    \centering
\caption{Applying our loss- and sampling-based N2V modifications when using Feature Propagation to extend embeddings.}
    \label{tab:fp+}
\begin{tabular}{lrr}
\toprule
model &  MLP & GraphSAGE \\
\midrule
FP  & $62.97$ & $63.20$ \\
FP w loss & $63.04$ & $63.22$ \\
FP w sampling & $62.98$ & $62.98$  \\
\bottomrule
\end{tabular}
\end{table}

\subsection{Ablation Study on the Post-hoc Method}
\label{sec:abl}

\begin{figure}[ht!]
    \centering
    \includegraphics[width=\linewidth]{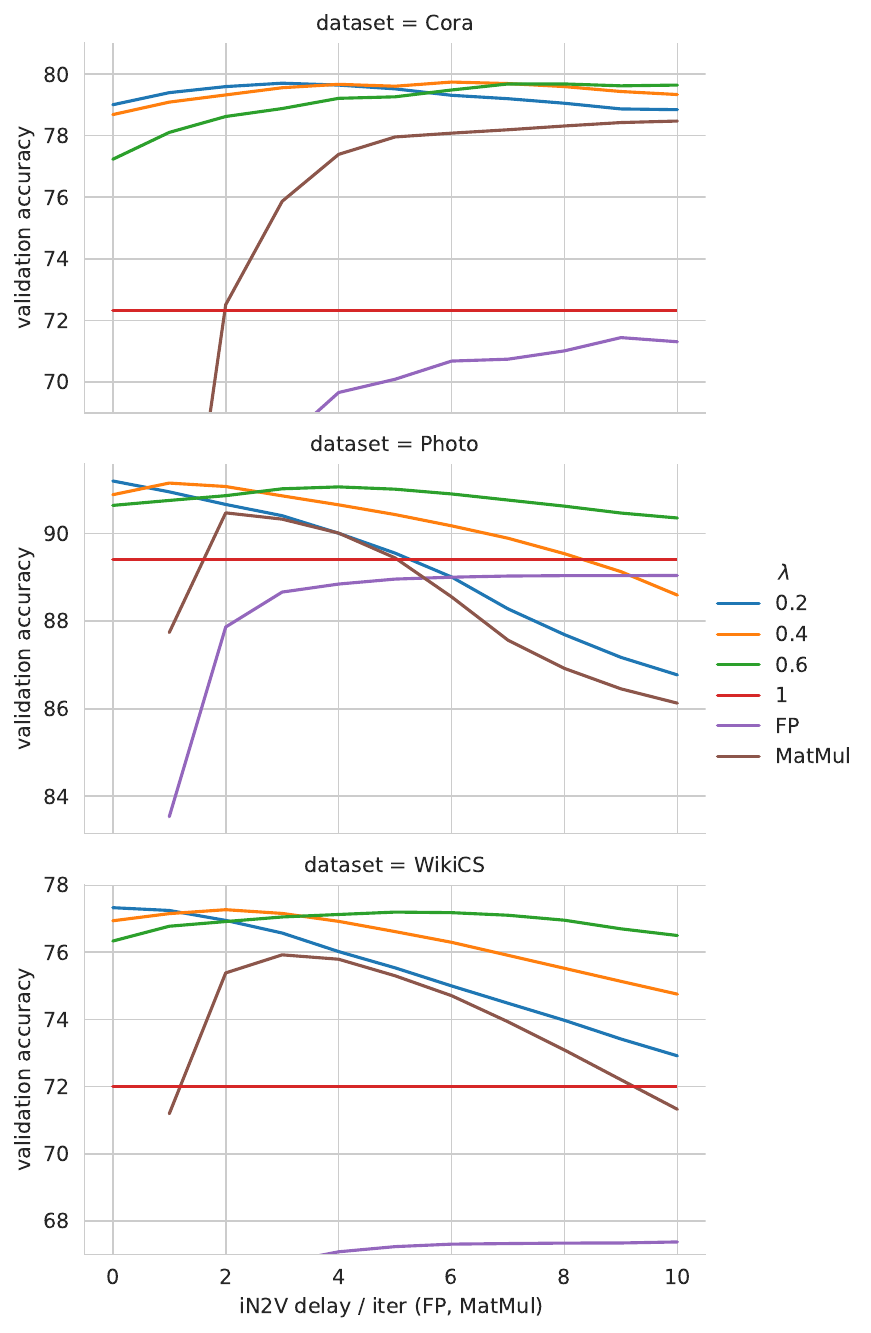}
    \caption{Ablation of the effect of $\lambda$ for different delays vs Feature Propagation and MatMul.}
    \label{fig:abl_lambda}
\end{figure}

We perform ablation studies on our post-hoc method and the loss modification.
For this, we use the $40\%$ training split, set the N2V hyperparameters $p$ and $q$ to $1$, embeddings size to $256$, and learning rate to $0.01$.
For a sensitivity analysis of these hyperparameters, see Appendix \ref{appx:hyper}.

Figure \ref{fig:abl_lambda} shows the effect of $\lambda$ and $delay$ in our post-hoc extension, Feature Propagation, as well as a MatMul baseline that multiplies the embeddings matrix $iter$-many times with the adjacency matrix.
When increasing $iter$ or $delay$, the post-hoc methods and MatMul increase to a maximum, which depends on $\lambda$, and then drop off again.
This nicely shows the trade-off we discuss in Section~\ref{sec:methods}. 
More iterations allow an adaption to new paths from the test split, but too many iterations lead to a convergence of all embeddings to a graph average.
For $\lambda = 1$, this is not the case as embeddings do not change once they are set, which means that the embedding and performance are fixed for $delay \geq 0$.
As Feature Propagation keeps the training embeddings fixed, its performance increases with more iterations.
Overall, we can observe that adapting training embeddings is important, as post-hoc with $\lambda < 1$ and MatMul outperform both Feature Propagation and post-hoc with $\lambda = 1$.

\begin{figure}[ht]
    \centering
    \subfigure[Cora]{
        \includegraphics[width=0.49\linewidth]{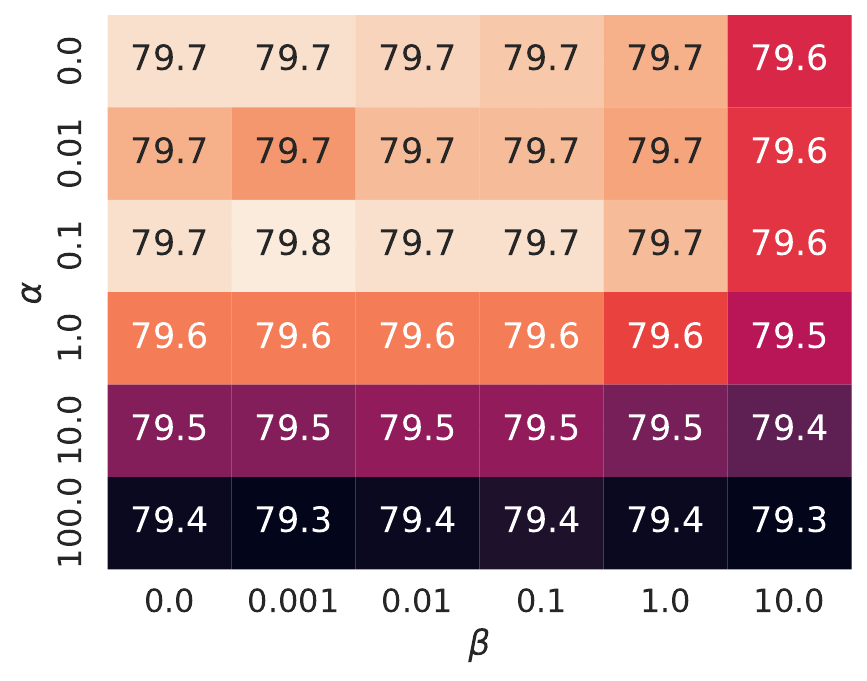}
    }\subfigure[Photo]{
        \includegraphics[width=0.49\linewidth]{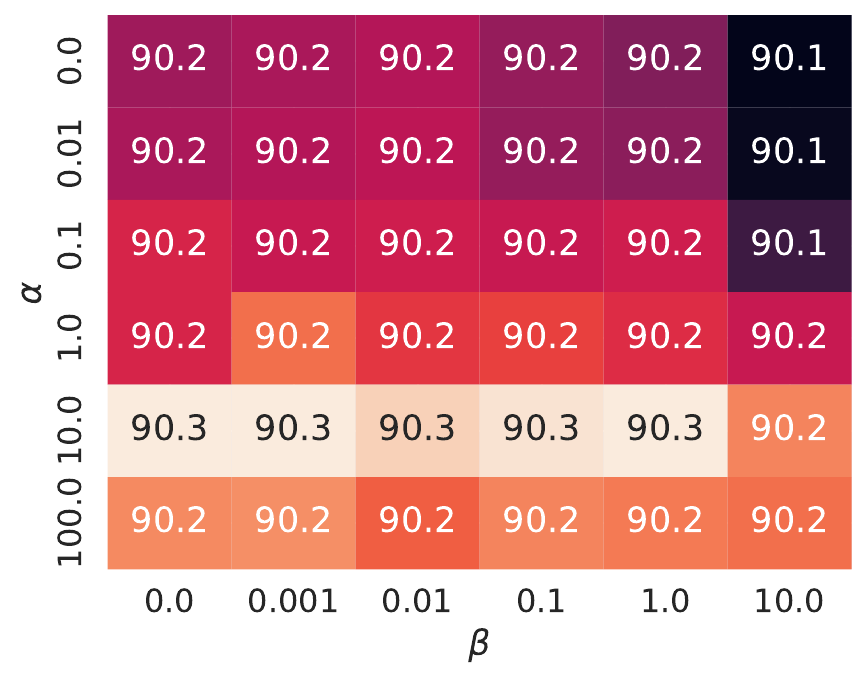}
    }
    \caption{Validation accuracy for different weights for $\alpha$ and $\beta$.}
    \label{fig:some_prep_loss}
\end{figure}

Figure~\ref{fig:some_prep_loss} shows the effect of the loss weights $\alpha$ and $\beta$ on logistic regression validation accuracy.
As we already saw in our main results, the loss-based N2V modification gives a small performance improvement.
The figure suggests that $\alpha$ has a bigger influence on the performance as long as $\beta$ is not too high.

\section{Limitations and Future Work}

Our iN2V algorithm uses N2V and shares some of the limitations.
N2V embeddings do not add any information for datasets where neighborhood structures alone are irrelevant.
We have observed this case with the Actor and Roman-Empire datasets, where the performance when using only N2V embeddings as model input was close to predicting the largest class.
Many molecule datasets consist of thousands of individual subgraphs, where the train and test split is done on a per-graph level.
In cases where there is no path between the training and test nodes, our post-hoc extensions (like Feature Propagation) cannot provide embeddings for test nodes in an inductive setting.
Our method generally can only provide embeddings to test nodes with a path to at least one train node.
If this affects only a few nodes, then iN2V still shows good performance.
We have shown this with Cora and Citeseer, which have $78$ and $438$ components, respectively.
Some of these components have no nodes in the training set when using the small training splits.

As our post-hoc extension is flexible and not tied to N2V, other shallow embedding methods can be used for datasets where the neighborhood structure does not provide helpful information.
For example, one could use struc2vec, which focuses on the similarity of neighborhood degree distributions, to obtain better embeddings for heterophilic datasets.
In this case, it might be better to utilize the neighborhood-similarity-based graph struc2vec builds for the post-hoc algorithm.
Another limitation and avenue for future work is that we used random sampling to create the dataset splits.
Using random splits is important for a fair evaluation of different embeddings and models~\cite{shchur}, but in the inductive case, this leads to some unconnected training nodes in the splits with few training nodes.
Especially for those splits, it might be beneficial to use a biased sampler that prefers nodes with edges to already sampled nodes to obtain better-connected training sets.

Our iN2V is not limited to simple graphs; it can already deal with multi-edges and self-loops.
Self-loops increase the influence of $h_v$ in Equation~\ref{1b} to more than $\lambda$ as it also appears in $m_{N_s(v)}$.
Edge weights could be incorporated into the mean neighbor embedding $m_{N_s(v)}$ by replacing the mean with a weighted sum using normalized edge weights.
For KGs, iN2V could be used as-is or by replacing N2V with a KG-focused embedding like TransE that embeds vertices and relations.
When doing so, Equations~\ref{1b} and \ref{1c} have to be adapted to incorporate the relation embeddings.

\section{Conclusion}
\label{sec:conclusion}

We introduced iN2V, a general post-hoc extension to induce embeddings to unseen nodes in the inductive setup.
We modified the training algorithm of N2V to obtain embeddings better suited to this induction.
Our extensive experiments on different datasets, training splits, and using different GNNs on the embeddings showed that iN2V works well and beats the comparable baselines.
For some datasets and splits, iN2V even outperforms using the original graph features.
In our detailed discussion and ablation, we have shown that our post-hoc extensions perform remarkably well. 
At the same time, the modifications of the N2V training have a smaller influence on the final performance.
Our experiments also showed general limitations of N2V-based approaches for some of the heterophilic datasets, where the performance remained close to random.

\section*{Acknowledgements}
This work was performed on the computational resource bwUniCluster funded by the Ministry of Science, Research and the Arts Baden-Württemberg and the Universities of the State of Baden-Württemberg, Germany, within the framework program bwHPC

\section*{Impact Statement}

This paper aims to transfer shallow embedding methods like N2V to the inductive setup, where only the graph induced by training nodes is known during training.
Our work can be used to apply GNNs to new nodes in the inductive setup, for example new users in social networks or new products in co-purchase graphs, and outperforms existing methods.
While our work has implications, particularly in improving the handling of nodes with missing features or in dealing with feature-less graphs, we believe that no specific societal consequences require immediate emphasis in this context.

\bibliography{example_paper}

\begin{thebibliography}{50}
\providecommand{\natexlab}[1]{#1}
\providecommand{\url}[1]{\texttt{#1}}
\expandafter\ifx\csname urlstyle\endcsname\relax
  \providecommand{\doi}[1]{doi: #1}\else
  \providecommand{\doi}{doi: \begingroup \urlstyle{rm}\Url}\fi

\bibitem[Abu{-}El{-}Haija et~al.(2019)Abu{-}El{-}Haija, Perozzi, Kapoor, Alipourfard, Lerman, Harutyunyan, Steeg, and Galstyan]{mixhop}
Abu{-}El{-}Haija, S., Perozzi, B., Kapoor, A., Alipourfard, N., Lerman, K., Harutyunyan, H., Steeg, G.~V., and Galstyan, A.
\newblock Mixhop: Higher-order graph convolutional architectures via sparsified neighborhood mixing.
\newblock In Chaudhuri, K. and Salakhutdinov, R. (eds.), \emph{Proceedings of the 36th International Conference on Machine Learning, {ICML} 2019, 9-15 June 2019, Long Beach, California, {USA}}, volume~97 of \emph{Proceedings of Machine Learning Research}, pp.\  21--29. {PMLR}, 2019.
\newblock URL \url{http://proceedings.mlr.press/v97/abu-el-haija19a.html}.

\bibitem[Adhikari et~al.(2018)Adhikari, Zhang, Ramakrishnan, and Prakash]{sub2vec}
Adhikari, B., Zhang, Y., Ramakrishnan, N., and Prakash, B.~A.
\newblock Sub2vec: Feature learning for subgraphs.
\newblock In Phung, D.~Q., Tseng, V.~S., Webb, G.~I., Ho, B., Ganji, M., and Rashidi, L. (eds.), \emph{Advances in Knowledge Discovery and Data Mining - 22nd Pacific-Asia Conference, {PAKDD} 2018, Melbourne, VIC, Australia, June 3-6, 2018, Proceedings, Part {II}}, volume 10938 of \emph{Lecture Notes in Computer Science}, pp.\  170--182. Springer, 2018.
\newblock \doi{10.1007/978-3-319-93037-4\_14}.
\newblock URL \url{https://doi.org/10.1007/978-3-319-93037-4\_14}.

\bibitem[Bi et~al.(2024)Bi, Du, Fu, Wang, Han, and Zhang]{hete_rew}
Bi, W., Du, L., Fu, Q., Wang, Y., Han, S., and Zhang, D.
\newblock Make heterophilic graphs better fit {GNN:} {A} graph rewiring approach.
\newblock \emph{{IEEE} Trans. Knowl. Data Eng.}, 36\penalty0 (12):\penalty0 8744--8757, 2024.
\newblock \doi{10.1109/TKDE.2024.3441766}.
\newblock URL \url{https://doi.org/10.1109/TKDE.2024.3441766}.

\bibitem[Bordes et~al.(2013)Bordes, Usunier, Garc{\'{\i}}a{-}Dur{\'{a}}n, Weston, and Yakhnenko]{transE}
Bordes, A., Usunier, N., Garc{\'{\i}}a{-}Dur{\'{a}}n, A., Weston, J., and Yakhnenko, O.
\newblock Translating embeddings for modeling multi-relational data.
\newblock In Burges, C. J.~C., Bottou, L., Ghahramani, Z., and Weinberger, K.~Q. (eds.), \emph{Advances in Neural Information Processing Systems 26: 27th Annual Conference on Neural Information Processing Systems 2013. Proceedings of a meeting held December 5-8, 2013, Lake Tahoe, Nevada, United States}, pp.\  2787--2795, 2013.
\newblock URL \url{https://proceedings.neurips.cc/paper/2013/hash/1cecc7a77928ca8133fa24680a88d2f9-Abstract.html}.

\bibitem[Chen et~al.(2020)Chen, Wei, Huang, Ding, and Li]{gcnii}
Chen, M., Wei, Z., Huang, Z., Ding, B., and Li, Y.
\newblock Simple and deep graph convolutional networks.
\newblock In \emph{Proceedings of the 37th International Conference on Machine Learning, {ICML} 2020, 13-18 July 2020, Virtual Event}, volume 119 of \emph{Proceedings of Machine Learning Research}, pp.\  1725--1735. {PMLR}, 2020.
\newblock URL \url{http://proceedings.mlr.press/v119/chen20v.html}.

\bibitem[Chen et~al.(2021)Chen, Zhang, Yuan, Jia, and Chen]{FedE}
Chen, M., Zhang, W., Yuan, Z., Jia, Y., and Chen, H.
\newblock Fede: Embedding knowledge graphs in federated setting.
\newblock In \emph{IJCKG'21: The 10th International Joint Conference on Knowledge Graphs, Virtual Event, Thailand, December 6 - 8, 2021}, pp.\  80--88. {ACM}, 2021.
\newblock \doi{10.1145/3502223.3502233}.
\newblock URL \url{https://doi.org/10.1145/3502223.3502233}.

\bibitem[Chen et~al.(2022)Chen, Mishra, Franceschi, Minervini, Stenetorp, and Riedel]{chen2022refactor}
Chen, Y., Mishra, P., Franceschi, L., Minervini, P., Stenetorp, P., and Riedel, S.
\newblock Refactor gnns: Revisiting factorisation-based models from a message-passing perspective.
\newblock In Koyejo, S., Mohamed, S., Agarwal, A., Belgrave, D., Cho, K., and Oh, A. (eds.), \emph{Advances in Neural Information Processing Systems 35: Annual Conference on Neural Information Processing Systems 2022, NeurIPS 2022, New Orleans, LA, USA, November 28 - December 9, 2022}, 2022.
\newblock URL \url{http://papers.nips.cc/paper\_files/paper/2022/hash/66f7a3df255c47b2e72f30b310a7e44a-Abstract-Conference.html}.

\bibitem[Chien et~al.(2021)Chien, Peng, Li, and Milenkovic]{prgnn}
Chien, E., Peng, J., Li, P., and Milenkovic, O.
\newblock Adaptive universal generalized pagerank graph neural network.
\newblock In \emph{9th International Conference on Learning Representations, {ICLR} 2021, Virtual Event, Austria, May 3-7, 2021}. OpenReview.net, 2021.
\newblock URL \url{https://openreview.net/forum?id=n6jl7fLxrP}.

\bibitem[Donnat et~al.(2018)Donnat, Zitnik, Hallac, and Leskovec]{graphwave}
Donnat, C., Zitnik, M., Hallac, D., and Leskovec, J.
\newblock Learning structural node embeddings via diffusion wavelets.
\newblock In Guo, Y. and Farooq, F. (eds.), \emph{Proceedings of the 24th {ACM} {SIGKDD} International Conference on Knowledge Discovery {\&} Data Mining, {KDD} 2018, London, UK, August 19-23, 2018}, pp.\  1320--1329. {ACM}, 2018.
\newblock \doi{10.1145/3219819.3220025}.
\newblock URL \url{https://doi.org/10.1145/3219819.3220025}.

\bibitem[Grover \& Leskovec(2016)Grover and Leskovec]{n2v}
Grover, A. and Leskovec, J.
\newblock node2vec: Scalable feature learning for networks.
\newblock In Krishnapuram, B., Shah, M., Smola, A.~J., Aggarwal, C.~C., Shen, D., and Rastogi, R. (eds.), \emph{Proceedings of the 22nd {ACM} {SIGKDD} International Conference on Knowledge Discovery and Data Mining, San Francisco, CA, USA, August 13-17, 2016}, pp.\  855--864. {ACM}, 2016.
\newblock \doi{10.1145/2939672.2939754}.
\newblock URL \url{https://doi.org/10.1145/2939672.2939754}.

\bibitem[Hahn \& Paulheim(2024)Hahn and Paulheim]{hahn2024rdf2vec}
Hahn, S.~H. and Paulheim, H.
\newblock Rdf2vec embeddings for updateable knowledge graphs--reuse, don’t retrain!
\newblock \emph{ESWC Posters and Demos}, 2024.

\bibitem[Hamilton et~al.(2017)Hamilton, Ying, and Leskovec]{sage}
Hamilton, W.~L., Ying, Z., and Leskovec, J.
\newblock Inductive representation learning on large graphs.
\newblock In Guyon, I., von Luxburg, U., Bengio, S., Wallach, H.~M., Fergus, R., Vishwanathan, S. V.~N., and Garnett, R. (eds.), \emph{Advances in Neural Information Processing Systems 30: Annual Conference on Neural Information Processing Systems 2017, December 4-9, 2017, Long Beach, CA, {USA}}, pp.\  1024--1034, 2017.
\newblock URL \url{https://proceedings.neurips.cc/paper/2017/hash/5dd9db5e033da9c6fb5ba83c7a7ebea9-Abstract.html}.

\bibitem[Hu et~al.(2021)Hu, You, Wang, Wang, Zhou, and Gao]{graph-mlp}
Hu, Y., You, H., Wang, Z., Wang, Z., Zhou, E., and Gao, Y.
\newblock Graph-mlp: Node classification without message passing in graph.
\newblock \emph{CoRR}, abs/2106.04051, 2021.
\newblock URL \url{https://arxiv.org/abs/2106.04051}.

\bibitem[Jia \& Benson(2022)Jia and Benson]{unifyingmodel}
Jia, J. and Benson, A.~R.
\newblock A unifying generative model for graph learning algorithms: Label propagation, graph convolutions, and combinations.
\newblock \emph{{SIAM} J. Math. Data Sci.}, 4\penalty0 (1):\penalty0 100--125, 2022.
\newblock \doi{10.1137/21M1395351}.
\newblock URL \url{https://doi.org/10.1137/21m1395351}.

\bibitem[Kipf \& Welling(2017)Kipf and Welling]{gcn}
Kipf, T.~N. and Welling, M.
\newblock Semi-supervised classification with graph convolutional networks.
\newblock In \emph{5th International Conference on Learning Representations, {ICLR} 2017, Toulon, France, April 24-26, 2017, Conference Track Proceedings}. OpenReview.net, 2017.
\newblock URL \url{https://openreview.net/forum?id=SJU4ayYgl}.

\bibitem[Kohn et~al.(2024)Kohn, Hoffmann, and Scherp]{esmlp}
Kohn, M., Hoffmann, M., and Scherp, A.
\newblock Edge-splitting {MLP}: Node classification on homophilic and heterophilic graphs without message passing.
\newblock In \emph{The Third Learning on Graphs Conference}, 2024.
\newblock URL \url{https://openreview.net/forum?id=BQEb4r21cm}.

\bibitem[Lacroix et~al.(2018)Lacroix, Usunier, and Obozinski]{canontensordecom}
Lacroix, T., Usunier, N., and Obozinski, G.
\newblock Canonical tensor decomposition for knowledge base completion.
\newblock In Dy, J.~G. and Krause, A. (eds.), \emph{Proceedings of the 35th International Conference on Machine Learning, {ICML} 2018, Stockholmsm{\"{a}}ssan, Stockholm, Sweden, July 10-15, 2018}, volume~80 of \emph{Proceedings of Machine Learning Research}, pp.\  2869--2878. {PMLR}, 2018.
\newblock URL \url{http://proceedings.mlr.press/v80/lacroix18a.html}.

\bibitem[Le \& Mikolov(2014)Le and Mikolov]{paragraph2v}
Le, Q.~V. and Mikolov, T.
\newblock Distributed representations of sentences and documents.
\newblock In \emph{Proceedings of the 31th International Conference on Machine Learning, {ICML} 2014, Beijing, China, 21-26 June 2014}, volume~32 of \emph{{JMLR} Workshop and Conference Proceedings}, pp.\  1188--1196. JMLR.org, 2014.
\newblock URL \url{http://proceedings.mlr.press/v32/le14.html}.

\bibitem[Lell \& Scherp(2024)Lell and Scherp]{hm}
Lell, N. and Scherp, A.
\newblock Hyperaggregation: Aggregating over graph edges with hypernetworks.
\newblock In \emph{International Joint Conference on Neural Networks, {IJCNN} 2024, Yokohama, Japan, June 30 - July 5, 2024}, pp.\  1--9. {IEEE}, 2024.
\newblock \doi{10.1109/IJCNN60899.2024.10650980}.
\newblock URL \url{https://doi.org/10.1109/IJCNN60899.2024.10650980}.

\bibitem[Lim et~al.(2021)Lim, Hohne, Li, Huang, Gupta, Bhalerao, and Lim]{linkx}
Lim, D., Hohne, F., Li, X., Huang, S.~L., Gupta, V., Bhalerao, O., and Lim, S.
\newblock Large scale learning on non-homophilous graphs: New benchmarks and strong simple methods.
\newblock In Ranzato, M., Beygelzimer, A., Dauphin, Y.~N., Liang, P., and Vaughan, J.~W. (eds.), \emph{Advances in Neural Information Processing Systems 34: Annual Conference on Neural Information Processing Systems 2021, NeurIPS 2021, December 6-14, 2021, virtual}, pp.\  20887--20902, 2021.
\newblock URL \url{https://proceedings.neurips.cc/paper/2021/hash/ae816a80e4c1c56caa2eb4e1819cbb2f-Abstract.html}.

\bibitem[Longa et~al.(2023)Longa, Lachi, Santin, Bianchini, Lepri, Lio, Scarselli, and Passerini]{tempgnns}
Longa, A., Lachi, V., Santin, G., Bianchini, M., Lepri, B., Lio, P., Scarselli, F., and Passerini, A.
\newblock Graph neural networks for temporal graphs: State of the art, open challenges, and opportunities.
\newblock \emph{Trans. Mach. Learn. Res.}, 2023, 2023.
\newblock URL \url{https://openreview.net/forum?id=pHCdMat0gI}.

\bibitem[Luan et~al.(2022)Luan, Hua, Lu, Zhu, Zhao, Zhang, Chang, and Precup]{acm}
Luan, S., Hua, C., Lu, Q., Zhu, J., Zhao, M., Zhang, S., Chang, X., and Precup, D.
\newblock Revisiting heterophily for graph neural networks.
\newblock In Koyejo, S., Mohamed, S., Agarwal, A., Belgrave, D., Cho, K., and Oh, A. (eds.), \emph{Advances in Neural Information Processing Systems 35: Annual Conference on Neural Information Processing Systems 2022, NeurIPS 2022, New Orleans, LA, USA, November 28 - December 9, 2022}, 2022.
\newblock URL \url{http://papers.nips.cc/paper\_files/paper/2022/hash/092359ce5cf60a80e882378944bf1be4-Abstract-Conference.html}.

\bibitem[Mernyei \& Cangea(2020)Mernyei and Cangea]{wikics}
Mernyei, P. and Cangea, C.
\newblock Wiki-cs: A wikipedia-based benchmark for graph neural networks.
\newblock \emph{arXiv preprint arXiv:2007.02901}, 2020.

\bibitem[Mikolov et~al.(2013)Mikolov, Chen, Corrado, and Dean]{Word2vec}
Mikolov, T., Chen, K., Corrado, G., and Dean, J.
\newblock Efficient estimation of word representations in vector space.
\newblock In Bengio, Y. and LeCun, Y. (eds.), \emph{1st International Conference on Learning Representations, {ICLR} 2013, Scottsdale, Arizona, USA, May 2-4, 2013, Workshop Track Proceedings}, 2013.
\newblock URL \url{http://arxiv.org/abs/1301.3781}.

\bibitem[Mironov \& Prokhorenkova(2024)Mironov and Prokhorenkova]{unbiased}
Mironov, M. and Prokhorenkova, L.
\newblock Revisiting graph homophily measures.
\newblock In \emph{The Third Learning on Graphs Conference}, 2024.
\newblock URL \url{https://openreview.net/forum?id=fiFBjLD0LV}.

\bibitem[Namata et~al.(2012)Namata, London, Getoor, Huang, and EDU]{pubmed}
Namata, G., London, B., Getoor, L., Huang, B., and EDU, U.
\newblock Query-driven active surveying for collective classification.
\newblock In \emph{10th International Workshop on Mining and Learning with Graphs}, 2012.

\bibitem[Narayanan et~al.(2016)Narayanan, Chandramohan, Chen, Liu, and Saminathan]{subgraph2vec}
Narayanan, A., Chandramohan, M., Chen, L., Liu, Y., and Saminathan, S.
\newblock subgraph2vec: Learning distributed representations of rooted sub-graphs from large graphs.
\newblock \emph{CoRR}, abs/1606.08928, 2016.
\newblock URL \url{http://arxiv.org/abs/1606.08928}.

\bibitem[Narayanan et~al.(2017)Narayanan, Chandramohan, Venkatesan, Chen, Liu, and Jaiswal]{graph2vec}
Narayanan, A., Chandramohan, M., Venkatesan, R., Chen, L., Liu, Y., and Jaiswal, S.
\newblock graph2vec: Learning distributed representations of graphs.
\newblock \emph{CoRR}, abs/1707.05005, 2017.
\newblock URL \url{http://arxiv.org/abs/1707.05005}.

\bibitem[Pei et~al.(2020)Pei, Wei, Chang, Lei, and Yang]{geomgcn}
Pei, H., Wei, B., Chang, K.~C., Lei, Y., and Yang, B.
\newblock Geom-gcn: Geometric graph convolutional networks.
\newblock In \emph{{ICLR} 2020}. OpenReview.net, 2020.
\newblock URL \url{https://openreview.net/forum?id=S1e2agrFvS}.

\bibitem[Perozzi et~al.(2014)Perozzi, Al{-}Rfou, and Skiena]{deepwalk}
Perozzi, B., Al{-}Rfou, R., and Skiena, S.
\newblock Deepwalk: online learning of social representations.
\newblock In Macskassy, S.~A., Perlich, C., Leskovec, J., Wang, W., and Ghani, R. (eds.), \emph{The 20th {ACM} {SIGKDD} International Conference on Knowledge Discovery and Data Mining, {KDD} '14, New York, NY, {USA} - August 24 - 27, 2014}, pp.\  701--710. {ACM}, 2014.
\newblock \doi{10.1145/2623330.2623732}.
\newblock URL \url{https://doi.org/10.1145/2623330.2623732}.

\bibitem[Platonov et~al.(2023{\natexlab{a}})Platonov, Kuznedelev, Babenko, and Prokhorenkova]{adj_hom}
Platonov, O., Kuznedelev, D., Babenko, A., and Prokhorenkova, L.
\newblock Characterizing graph datasets for node classification: Homophily-heterophily dichotomy and beyond.
\newblock In Oh, A., Naumann, T., Globerson, A., Saenko, K., Hardt, M., and Levine, S. (eds.), \emph{Advances in Neural Information Processing Systems 36: Annual Conference on Neural Information Processing Systems 2023, NeurIPS 2023, New Orleans, LA, USA, December 10 - 16, 2023}, 2023{\natexlab{a}}.
\newblock URL \url{http://papers.nips.cc/paper\_files/paper/2023/hash/01b681025fdbda8e935a66cc5bb6e9de-Abstract-Conference.html}.

\bibitem[Platonov et~al.(2023{\natexlab{b}})Platonov, Kuznedelev, Diskin, Babenko, and Prokhorenkova]{heter}
Platonov, O., Kuznedelev, D., Diskin, M., Babenko, A., and Prokhorenkova, L.
\newblock A critical look at the evaluation of gnns under heterophily: Are we really making progress?
\newblock In \emph{{ICLR} 2023}. OpenReview.net, 2023{\natexlab{b}}.
\newblock URL \url{https://openreview.net/pdf?id=tJbbQfw-5wv}.

\bibitem[Polleres et~al.(2023)Polleres, Pernisch, Bonifati, Dell'Aglio, Dobriy, Dumbrava, Etcheverry, Ferranti, Hose, Jim\'{e}nez-Ruiz, Lissandrini, Scherp, Tommasini, and Wachs]{TGDK}
Polleres, A., Pernisch, R., Bonifati, A., Dell'Aglio, D., Dobriy, D., Dumbrava, S., Etcheverry, L., Ferranti, N., Hose, K., Jim\'{e}nez-Ruiz, E., Lissandrini, M., Scherp, A., Tommasini, R., and Wachs, J.
\newblock {How Does Knowledge Evolve in Open Knowledge Graphs?}
\newblock \emph{Transactions on Graph Data and Knowledge}, 1\penalty0 (1):\penalty0 11:1--11:59, 2023.
\newblock \doi{10.4230/TGDK.1.1.11}.
\newblock URL \url{https://drops.dagstuhl.de/entities/document/10.4230/TGDK.1.1.11}.

\bibitem[Ribeiro et~al.(2017)Ribeiro, Saverese, and Figueiredo]{struc2vec}
Ribeiro, L. F.~R., Saverese, P. H.~P., and Figueiredo, D.~R.
\newblock \emph{struc2vec}: Learning node representations from structural identity.
\newblock In \emph{Proceedings of the 23rd {ACM} {SIGKDD} International Conference on Knowledge Discovery and Data Mining, Halifax, NS, Canada, August 13 - 17, 2017}, pp.\  385--394. {ACM}, 2017.
\newblock \doi{10.1145/3097983.3098061}.
\newblock URL \url{https://doi.org/10.1145/3097983.3098061}.

\bibitem[Ristoski \& Paulheim(2016)Ristoski and Paulheim]{RDF2vec}
Ristoski, P. and Paulheim, H.
\newblock Rdf2vec: {RDF} graph embeddings for data mining.
\newblock In Groth, P., Simperl, E., Gray, A. J.~G., Sabou, M., Kr{\"{o}}tzsch, M., L{\'{e}}cu{\'{e}}, F., Fl{\"{o}}ck, F., and Gil, Y. (eds.), \emph{The Semantic Web - {ISWC} 2016 - 15th International Semantic Web Conference, Kobe, Japan, October 17-21, 2016, Proceedings, Part {I}}, volume 9981 of \emph{Lecture Notes in Computer Science}, pp.\  498--514, 2016.
\newblock \doi{10.1007/978-3-319-46523-4\_30}.
\newblock URL \url{https://doi.org/10.1007/978-3-319-46523-4\_30}.

\bibitem[Rossi et~al.(2022)Rossi, Kenlay, Gorinova, Chamberlain, Dong, and Bronstein]{featureprop}
Rossi, E., Kenlay, H., Gorinova, M.~I., Chamberlain, B.~P., Dong, X., and Bronstein, M.~M.
\newblock On the unreasonable effectiveness of feature propagation in learning on graphs with missing node features.
\newblock In Rieck, B. and Pascanu, R. (eds.), \emph{Learning on Graphs Conference, LoG 2022, 9-12 December 2022, Virtual Event}, volume 198 of \emph{Proceedings of Machine Learning Research}, pp.\ ~11. {PMLR}, 2022.
\newblock URL \url{https://proceedings.mlr.press/v198/rossi22a.html}.

\bibitem[Sancak et~al.(2024)Sancak, Balin, and Catalyurek]{nasc}
Sancak, K., Balin, M.~F., and Catalyurek, U.
\newblock Do we really need complicated graph learning models? -- a simple but effective baseline.
\newblock In \emph{The Third Learning on Graphs Conference}, 2024.
\newblock URL \url{https://openreview.net/forum?id=0664MgKEVz}.

\bibitem[Sen et~al.(2008)Sen, Namata, Bilgic, Getoor, Gallagher, and Eliassi{-}Rad]{CoraCiteseer}
Sen, P., Namata, G., Bilgic, M., Getoor, L., Gallagher, B., and Eliassi{-}Rad, T.
\newblock Collective classification in network data.
\newblock \emph{{AI} Mag.}, 2008.
\newblock \doi{10.1609/aimag.v29i3.2157}.

\bibitem[Shchur et~al.(2018)Shchur, Mumme, Bojchevski, and G{\"{u}}nnemann]{shchur}
Shchur, O., Mumme, M., Bojchevski, A., and G{\"{u}}nnemann, S.
\newblock Pitfalls of graph neural network evaluation.
\newblock \emph{CoRR}, 2018.
\newblock URL \url{http://arxiv.org/abs/1811.05868}.

\bibitem[Sun et~al.(2019)Sun, Deng, Nie, and Tang]{rotatE}
Sun, Z., Deng, Z., Nie, J., and Tang, J.
\newblock Rotate: Knowledge graph embedding by relational rotation in complex space.
\newblock In \emph{7th International Conference on Learning Representations, {ICLR} 2019, New Orleans, LA, USA, May 6-9, 2019}. OpenReview.net, 2019.
\newblock URL \url{https://openreview.net/forum?id=HkgEQnRqYQ}.

\bibitem[Tang et~al.(2015)Tang, Qu, Wang, Zhang, Yan, and Mei]{line}
Tang, J., Qu, M., Wang, M., Zhang, M., Yan, J., and Mei, Q.
\newblock {LINE:} large-scale information network embedding.
\newblock In Gangemi, A., Leonardi, S., and Panconesi, A. (eds.), \emph{Proceedings of the 24th International Conference on World Wide Web, {WWW} 2015, Florence, Italy, May 18-22, 2015}, pp.\  1067--1077. {ACM}, 2015.
\newblock \doi{10.1145/2736277.2741093}.
\newblock URL \url{https://doi.org/10.1145/2736277.2741093}.

\bibitem[Tian et~al.(2023)Tian, Zhang, Guo, Zhang, and Chawla]{nosmog}
Tian, Y., Zhang, C., Guo, Z., Zhang, X., and Chawla, N.~V.
\newblock Learning mlps on graphs: {A} unified view of effectiveness, robustness, and efficiency.
\newblock In \emph{The Eleventh International Conference on Learning Representations, {ICLR} 2023, Kigali, Rwanda, May 1-5, 2023}. OpenReview.net, 2023.
\newblock URL \url{https://openreview.net/forum?id=Cs3r5KLdoj}.

\bibitem[Trouillon et~al.(2016)Trouillon, Welbl, Riedel, Gaussier, and Bouchard]{complEx}
Trouillon, T., Welbl, J., Riedel, S., Gaussier, {\'{E}}., and Bouchard, G.
\newblock Complex embeddings for simple link prediction.
\newblock In Balcan, M. and Weinberger, K.~Q. (eds.), \emph{Proceedings of the 33nd International Conference on Machine Learning, {ICML} 2016, New York City, NY, USA, June 19-24, 2016}, volume~48 of \emph{{JMLR} Workshop and Conference Proceedings}, pp.\  2071--2080. JMLR.org, 2016.
\newblock URL \url{http://proceedings.mlr.press/v48/trouillon16.html}.

\bibitem[Velickovic et~al.(2018)Velickovic, Cucurull, Casanova, Romero, Li{\`{o}}, and Bengio]{gat}
Velickovic, P., Cucurull, G., Casanova, A., Romero, A., Li{\`{o}}, P., and Bengio, Y.
\newblock Graph attention networks.
\newblock In \emph{6th International Conference on Learning Representations, {ICLR} 2018, Vancouver, BC, Canada, April 30 - May 3, 2018, Conference Track Proceedings}. OpenReview.net, 2018.
\newblock URL \url{https://openreview.net/forum?id=rJXMpikCZ}.

\bibitem[Wang et~al.(2014)Wang, Zhang, Feng, and Chen]{transH}
Wang, Z., Zhang, J., Feng, J., and Chen, Z.
\newblock Knowledge graph embedding by translating on hyperplanes.
\newblock In Brodley, C.~E. and Stone, P. (eds.), \emph{Proceedings of the Twenty-Eighth {AAAI} Conference on Artificial Intelligence, July 27 -31, 2014, Qu{\'{e}}bec City, Qu{\'{e}}bec, Canada}, pp.\  1112--1119. {AAAI} Press, 2014.
\newblock \doi{10.1609/AAAI.V28I1.8870}.
\newblock URL \url{https://doi.org/10.1609/aaai.v28i1.8870}.

\bibitem[Wu et~al.(2019)Wu, Jr., Zhang, Fifty, Yu, and Weinberger]{sgc}
Wu, F., Jr., A. H.~S., Zhang, T., Fifty, C., Yu, T., and Weinberger, K.~Q.
\newblock Simplifying graph convolutional networks.
\newblock In Chaudhuri, K. and Salakhutdinov, R. (eds.), \emph{Proceedings of the 36th International Conference on Machine Learning, {ICML} 2019, 9-15 June 2019, Long Beach, California, {USA}}, volume~97 of \emph{Proceedings of Machine Learning Research}, pp.\  6861--6871. {PMLR}, 2019.
\newblock URL \url{http://proceedings.mlr.press/v97/wu19e.html}.

\bibitem[Xu et~al.(2018)Xu, Li, Tian, Sonobe, Kawarabayashi, and Jegelka]{jk}
Xu, K., Li, C., Tian, Y., Sonobe, T., Kawarabayashi, K., and Jegelka, S.
\newblock Representation learning on graphs with jumping knowledge networks.
\newblock In Dy, J.~G. and Krause, A. (eds.), \emph{Proceedings of the 35th International Conference on Machine Learning, {ICML} 2018, Stockholmsm{\"{a}}ssan, Stockholm, Sweden, July 10-15, 2018}, volume~80 of \emph{Proceedings of Machine Learning Research}, pp.\  5449--5458. {PMLR}, 2018.
\newblock URL \url{http://proceedings.mlr.press/v80/xu18c.html}.

\bibitem[Xu et~al.(2019)Xu, Hu, Leskovec, and Jegelka]{gin}
Xu, K., Hu, W., Leskovec, J., and Jegelka, S.
\newblock How powerful are graph neural networks?
\newblock In \emph{7th International Conference on Learning Representations, {ICLR} 2019, New Orleans, LA, USA, May 6-9, 2019}. OpenReview.net, 2019.
\newblock URL \url{https://openreview.net/forum?id=ryGs6iA5Km}.

\bibitem[Zhang et~al.(2022)Zhang, Liu, Sun, and Shah]{glnn}
Zhang, S., Liu, Y., Sun, Y., and Shah, N.
\newblock Graph-less neural networks: Teaching old mlps new tricks via distillation.
\newblock In \emph{The Tenth International Conference on Learning Representations, {ICLR} 2022, Virtual Event, April 25-29, 2022}. OpenReview.net, 2022.
\newblock URL \url{https://openreview.net/forum?id=4p6\_5HBWPCw}.

\bibitem[Zhu et~al.(2020)Zhu, Yan, Zhao, Heimann, Akoglu, and Koutra]{h2gcn}
Zhu, J., Yan, Y., Zhao, L., Heimann, M., Akoglu, L., and Koutra, D.
\newblock Beyond homophily in graph neural networks: Current limitations and effective designs.
\newblock In Larochelle, H., Ranzato, M., Hadsell, R., Balcan, M., and Lin, H. (eds.), \emph{Advances in Neural Information Processing Systems 33: Annual Conference on Neural Information Processing Systems 2020, NeurIPS 2020, December 6-12, 2020, virtual}, 2020.
\newblock URL \url{https://proceedings.neurips.cc/paper/2020/hash/58ae23d878a47004366189884c2f8440-Abstract.html}.

\end{thebibliography}
\bibliographystyle{icml2025}

\newpage
\appendix

\onecolumn 

\section*{Supplemental Materials}

\section{Hyperparameter Sensitivity}
\label{appx:hyper}
\begin{figure}[ht]
    \centering
    \subfigure[Cora]{
        \includegraphics[width=0.4\linewidth]{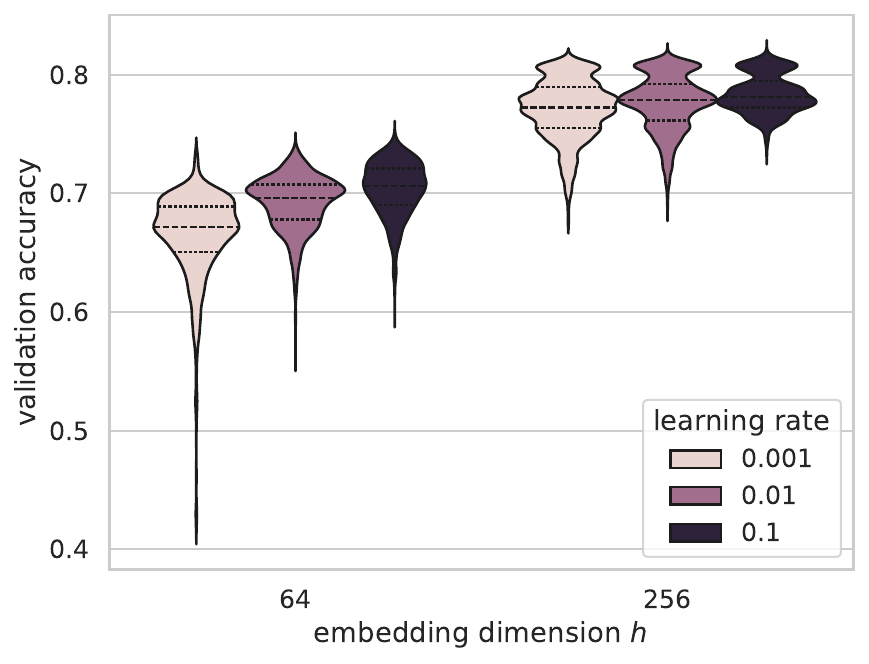}
        \label{fig:apx_Cora_h1}
    }
    \subfigure[Photo]{
        \includegraphics[width=0.4\linewidth]{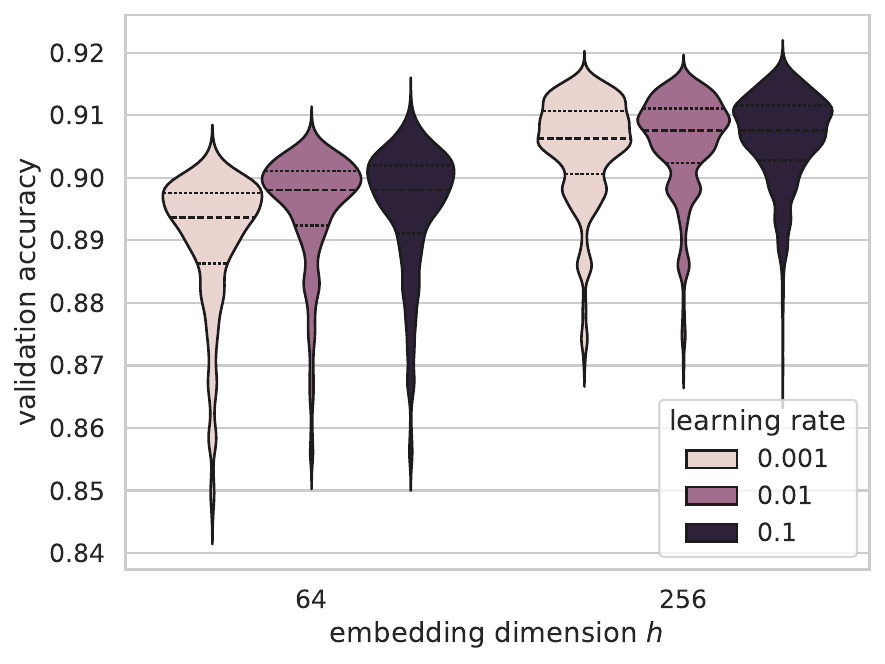}
        \label{fig:apx_Pho_h1}
    }\subfigure[WikiCS]{
        \includegraphics[width=0.4\linewidth]{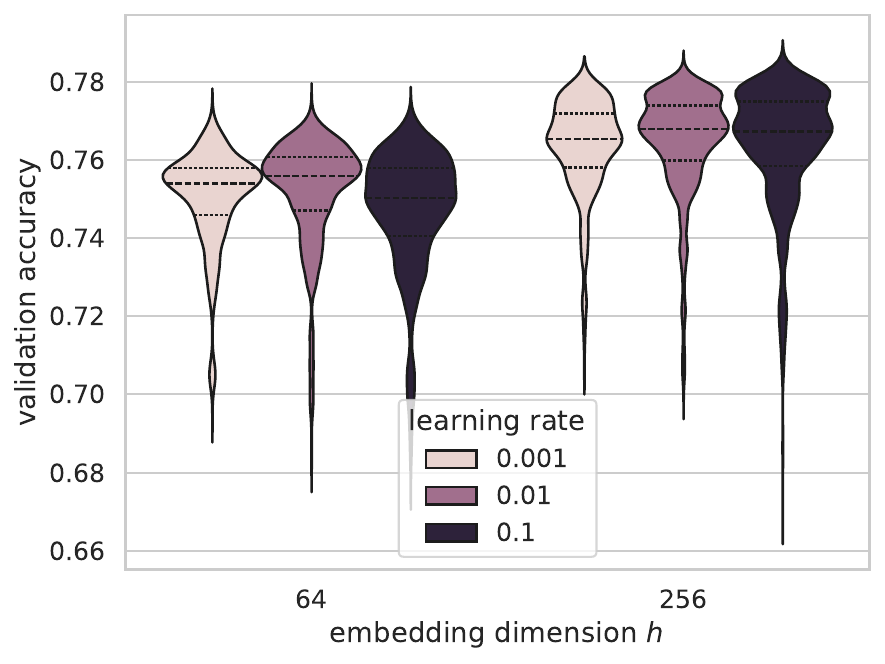}
        \label{fig:apx_WCS_h1}
    }
    \caption{Sensitivity to learning rate and embedding dimension.}
    \label{fig:apx_dim_lr}
\end{figure}

\begin{figure}[ht]
    \centering
    \subfigure[Cora]{
        \includegraphics[width=0.4\linewidth]{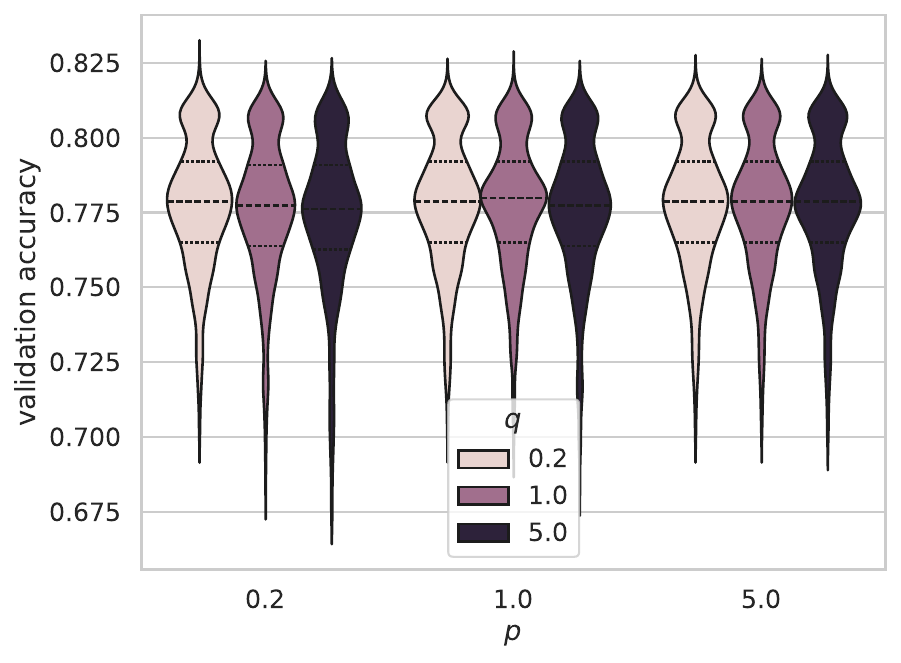}
        \label{fig:apx_Cora_h2}
    }
    \subfigure[Photo]{
        \includegraphics[width=0.4\linewidth]{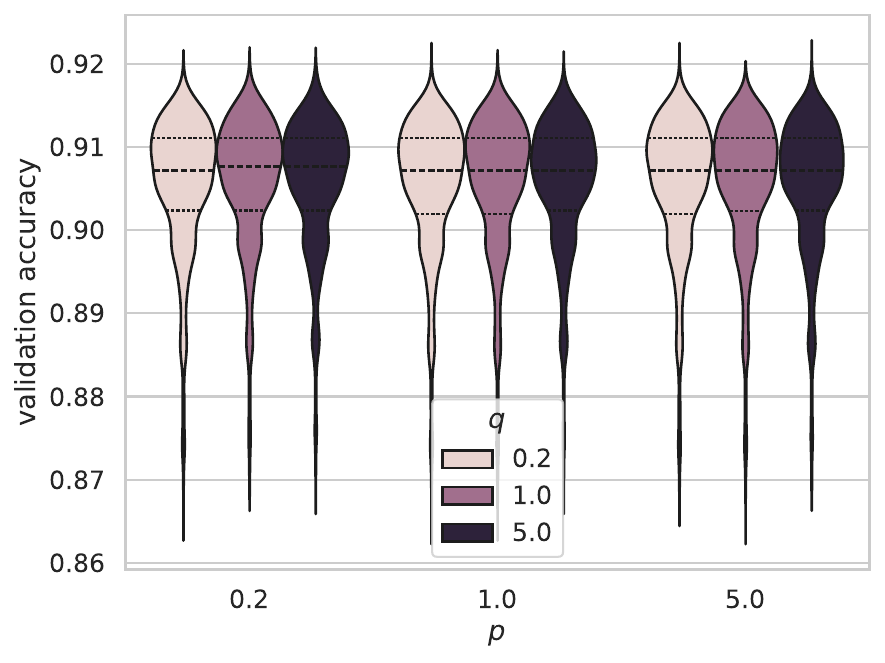}
        \label{fig:apx_Pho_h2}
    }\subfigure[WikiCS]{
        \includegraphics[width=0.4\linewidth]{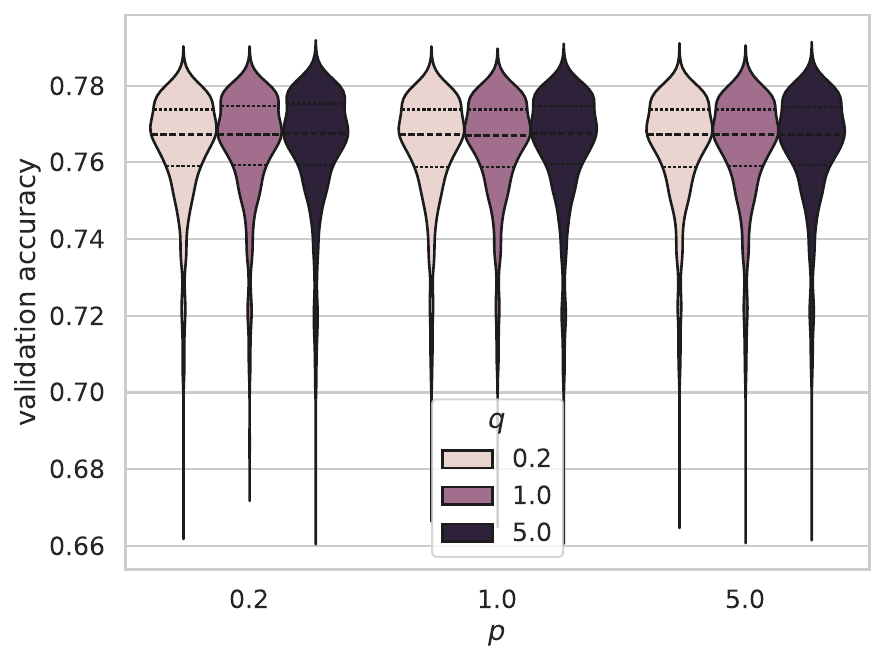}
        \label{fig:apx_WCS_h2}
    }
    \caption{Sensitivity to the N2V $p$ and $q$ hyperparameters.}
    \label{fig:apx_pq}
\end{figure}

This section follows up on Section \ref{sec:abl}.
Figure~\ref{fig:apx_dim_lr} shows the sensitivity of validation accuracy to the iN2V learning rate and embedding dimension while all other hyperparameters (including $\lambda$, $\alpha$, $\beta$, and $r$) are aggregated in the densities.
We can see that the embedding dimension has the biggest influence on the performance, especially for the Cora dataset.
For Figure~\ref{fig:apx_pq}, which shows the sensitivity to $p$ and $q$, we therefore fix the embedding dimension to $256$. 

\section{Complete Result Tables}
\label{appendix:res_full}

Tables~\ref{tab:full_mlp_base} and \ref{tab:full_sage_base} show the comparison of MLP and GraphSAGE on the iN2V embeddings vs baselines.
N2V, applied in the inductive setting, and Feature Propagation are comparable because they have access to the same information during training, while original features and N2V in the transductive setting utilize more information (the original features and all nodes during training).
Tables~\ref{tab:full_mlp_setups} and \ref{tab:full_sage_setups} show the MLP and GraphSAGE results when using different iN2V settings.

\section{Combining Original Graph Features with Trained N2V Embeddings}
\label{appx:cat}

As shown by related work, shallow embeddings like DeepWalk or N2V can be combined with the original graph features to improve the performance of GNNs.
This is especially helpful for MLPs who do not have access to structure information when only using the default graph features \cite{nosmog}.
Tables~\ref{tab:cat_mlp_base} and \ref{tab:cat_sage_base} compare the iN2V embeddings vs baselines with the original graph features concatenated to the input embeddings.
Tables~\ref{tab:cat_mlp_n2v} and \ref{tab:cat_sage_n2v} compare the different iN2V setups when concatenating the embeddings with the original graph features.

\section{Using other GNNs}
\label{appx:gat_gin}
The effectiveness of iN2V is not limited by the chosen GNN.
To demonstrate this, we additionally performed all experiments with GAT~\cite{gat}, see Tables~\ref{tab:full_gat_base}, \ref{tab:full_gat_setups}, \ref{tab:cat_gat_base}, and \ref{tab:cat_gat_n2v} and with GIN~\cite{gin}, see Tables~\ref{tab:full_gin_base}, \ref{tab:full_gin_setups}, \ref{tab:cat_gin_base}, and \ref{tab:cat_gin_n2v}.
For GAT, we set the number of attention heads to 8.
Otherwise, we used the same hyperparameter tuning procedure for both GAT and GIN as for GraphSAGE.

The results for GIN and GAT are in line with the GraphSAGE results.
iN2V outperforms all baselines for most splits on all datasets except Actor, where all N2V-based results amount to guessing the largest class.
Averaging the iN2V results over all datasets, splits, and whether to concatenate the iN2V embedding with the original graph features, MLP remains the best model with an average accuracy of $69.37$ points.
GraphSAGE reaches an average of $69.25$ points, GIN of $69.08$ points, and GAT of $68.64$ points.
This shows that an MLP can outperform classical GNNs without distillation or contrastive learning if enough structure information is provided with the features.

When considering all four GNNs, iN2V outperforms FP by $0.8$ points on homophilic and $0.3$ points on heterophilic datasets.
iN2V outperforms FP by $1.3$ points when using MLP as a classification model and by $0.4$ points when using message-passing GNNs.
iN2V outperforms FP by $0.9$ points when using only trained embeddings and by $0.3$ points when using both trained embeddings and the original graph features as input.
Finally, for the $10\%$ and $20\%$ training splits, iN2V leads over FP by $0.9$ points in contrast to the $0.4$ point lead for the $60\%$ and $80\%$ training splits.

\twocolumn

\begin{table}[]
\centering
\caption{Comparison of best iN2V variant vs baselines; MLP accuracy. Gray numbers are not directly comparable as they use additional information (graph features/transductive setup).}
\resizebox{\linewidth}{!}{
\begin{tabular}{lrrrrr} \toprule
& \multicolumn{5}{l}{Percentage of training data} \\ \midrule
 Dataset & 10\% & 20\% & 40\% & 60\% & 80\% \\ \midrule
\multicolumn{6}{l}{Cora} \\ \hline
N2V (inductive) & $30.04_{1.06}$ & $30.07_{1.15}$ & $30.12_{1.05}$ & $29.67_{1.04}$ & $30.70_{2.14}$ \\
Feature Propagation & $78.54_{1.42}$ & $81.11_{1.17}$ & $82.38_{1.01}$ & $83.76_{1.00}$ & $85.13_{2.27}$ \\
\textbf{iN2V} (own) & $\mathbf{79.84}_{1.27}$ & $\mathbf{81.70}_{1.19}$ & $\mathbf{83.93}_{1.16}$ & $\mathbf{84.69}_{0.95}$ & $\mathbf{85.57}_{2.22}$ \\ \hline
\textcolor{gray}{Original features} & \textcolor{gray}{$66.98_{1.70}$} & \textcolor{gray}{$71.34_{1.79}$} & \textcolor{gray}{$75.94_{1.38}$} & \textcolor{gray}{$78.78_{1.11}$} & \textcolor{gray}{$79.23_{1.95}$} \\
\textcolor{gray}{N2V (transductive)} & \textcolor{gray}{$78.44_{1.15}$} & \textcolor{gray}{$80.84_{1.49}$} & \textcolor{gray}{$82.82_{1.10}$} & \textcolor{gray}{$83.75_{1.20}$} & \textcolor{gray}{$83.10_{1.86}$} \\
\bottomrule
\end{tabular}}
\label{tab:full_mlp_base}
\end{table}

\begin{table}[]
\centering
\caption{Comparison of best iN2V variant vs baselines; GraphSAGE accuracy. Gray numbers are not directly comparable as they use additional information (graph features/transductive setup).}
\resizebox{\linewidth}{!}{
\begin{tabular}{lrrrrr} \toprule
& \multicolumn{5}{l}{Percentage of training data} \\ \midrule
 Dataset & 10\% & 20\% & 40\% & 60\% & 80\% \\ \midrule
\multicolumn{6}{l}{Cora} \\ \hline
N2V (inductive) & $42.18_{3.52}$ & $59.91_{4.15}$ & $75.07_{1.92}$ & $81.31_{1.56}$ & $\mathbf{84.50}_{1.44}$ \\
Feature Propagation & $77.91_{2.62}$ & $79.48_{2.19}$ & $81.03_{1.85}$ & $\mathbf{84.00}_{1.00}$ & $84.13_{2.35}$ \\
\textbf{iN2V} (own) & $\mathbf{78.88}_{1.45}$ & $\mathbf{80.94}_{1.58}$ & $\mathbf{83.30}_{1.09}$ & $83.73_{1.13}$ & $84.46_{2.08}$ \\ \hline
\textcolor{gray}{Original features} & \textcolor{gray}{$75.27_{2.63}$} & \textcolor{gray}{$83.37_{1.17}$} & \textcolor{gray}{$86.23_{1.77}$} & \textcolor{gray}{$86.99_{0.86}$} & \textcolor{gray}{$87.05_{1.20}$} \\
\textcolor{gray}{N2V (transductive)} & \textcolor{gray}{$79.25_{1.45}$} & \textcolor{gray}{$81.66_{1.29}$} & \textcolor{gray}{$83.81_{0.95}$} & \textcolor{gray}{$85.24_{0.83}$} & \textcolor{gray}{$86.01_{1.99}$} \\
\bottomrule
\end{tabular}}
\label{tab:full_sage_base}
\end{table}

\begin{table}[]
\centering
\caption{Comparison of N2V vs different iN2V setups, MLP accuracy.}
\resizebox{\linewidth}{!}{
\begin{tabular}{lrrrrr} \toprule
& \multicolumn{5}{l}{Percentage of training data} \\ \midrule
 Dataset & 10\% & 20\% & 40\% & 60\% & 80\% \\ \midrule
\multicolumn{6}{l}{Cora} \\ \hline
N2V (inductive) & $30.04_{1.06}$ & $30.07_{1.15}$ & $30.12_{1.05}$ & $29.67_{1.04}$ & $30.70_{2.14}$ \\
frozen ($\lambda = 1$) & $74.64_{1.75}$ & $76.71_{1.60}$ & $79.57_{1.24}$ & $81.75_{1.63}$ & $83.51_{1.79}$ \\
post-hoc & $\mathbf{79.84}_{1.27}$ & $80.28_{1.80}$ & $83.93_{1.16}$ & $84.00_{1.03}$ & $\mathbf{85.57}_{2.46}$ \\
p-h w losses & $79.38_{1.42}$ & $81.07_{1.58}$ & $\mathbf{83.99}_{1.06}$ & $83.75_{1.51}$ & $85.24_{2.33}$ \\
p-h w sampling & $79.57_{1.34}$ & $\mathbf{81.70}_{1.19}$ & $82.71_{0.84}$ & $\mathbf{84.69}_{0.95}$ & $\mathbf{85.57}_{2.22}$ \\
\bottomrule
    \end{tabular}}
    \label{tab:full_mlp_setups}
\end{table}

\begin{table}[]
\centering
\caption{Comparison of N2V vs different iN2V setups, GraphSAGE accuracy.}
\resizebox{\linewidth}{!}{
\begin{tabular}{lrrrrr} \toprule
& \multicolumn{5}{l}{Percentage of training data} \\ \midrule
 Dataset & 10\% & 20\% & 40\% & 60\% & 80\% \\ \midrule
\multicolumn{6}{l}{Cora} \\ \hline
N2V (inductive) & $42.18_{3.52}$ & $59.91_{4.15}$ & $75.07_{1.92}$ & $81.31_{1.56}$ & $84.50_{1.44}$ \\
frozen ($\lambda = 1$) & $74.76_{2.08}$ & $76.86_{1.66}$ & $80.75_{1.17}$ & $82.45_{2.14}$ & $84.13_{1.99}$ \\
post-hoc & $\mathbf{79.20}_{1.80}$ & $79.44_{1.55}$ & $\mathbf{83.45}_{1.08}$ & $\mathbf{83.73}_{1.13}$ & $84.46_{2.08}$ \\
p-h w losses & $78.88_{1.45}$ & $80.50_{1.26}$ & $83.30_{1.09}$ & $83.39_{1.74}$ & $\mathbf{84.69}_{1.65}$ \\
p-h w sampling & $79.11_{1.77}$ & $\mathbf{80.94}_{1.58}$ & $82.02_{1.34}$ & $83.43_{1.50}$ & $84.02_{1.88}$ \\
\bottomrule
    \end{tabular}}
    \label{tab:full_sage_setups}
\end{table}

\begin{table}[]
\centering
\caption{Comparison of best iN2V variant vs baselines concatenated with the original graph features; MLP accuracy. Gray numbers are not directly comparable as they use additional information (transductive setup).}
\resizebox{\linewidth}{!}{
\begin{tabular}{lrrrrr} \toprule
& \multicolumn{5}{l}{Percentage of training data} \\ \midrule
 Dataset & 10\% & 20\% & 40\% & 60\% & 80\% \\ \midrule
\multicolumn{6}{l}{Cora} \\ \hline
N2V (inductive) & $66.48_{1.68}$ & $70.54_{1.20}$ & $75.47_{1.91}$ & $76.09_{2.08}$ & $76.27_{1.68}$ \\
Feature Propagation & $\mathbf{82.24}_{1.06}$ & $83.49_{1.26}$ & $83.83_{1.49}$ & $85.59_{1.24}$ & $85.54_{1.48}$ \\
\textbf{iN2V} (own) & $82.01_{1.38}$ & $\mathbf{83.97}_{1.35}$ & $\mathbf{85.83}_{1.45}$ & $\mathbf{85.81}_{1.77}$ & $\mathbf{86.49}_{1.62}$ \\ 
Original features & $66.98_{1.70}$ & $71.34_{1.79}$ & $75.94_{1.38}$ & $78.78_{1.11}$ & $79.23_{1.95}$ \\ \hline
\textcolor{gray}{N2V (transductive)} & \textcolor{gray}{$81.94_{1.39}$} & \textcolor{gray}{$83.52_{1.13}$} & \textcolor{gray}{$85.99_{1.10}$} & \textcolor{gray}{$86.99_{1.44}$} & \textcolor{gray}{$86.13_{1.69}$} \\
\bottomrule
\end{tabular}}
\label{tab:cat_mlp_base}
\end{table}

\begin{table}[]
\centering
\caption{Comparison of best iN2V variant vs baselines concatenated with the original graph features; GraphSAGE accuracy. Gray numbers are not directly comparable as they use additional information (transductive setup).}
\resizebox{\linewidth}{!}{
\begin{tabular}{lrrrrr} \toprule
& \multicolumn{5}{l}{Percentage of training data} \\ \midrule
 Dataset & 10\% & 20\% & 40\% & 60\% & 80\% \\ \midrule
\multicolumn{6}{l}{Cora} \\ \hline
N2V (inductive) & $73.97_{3.35}$ & $81.92_{1.84}$ & $85.97_{1.67}$ & $86.97_{0.96}$ & $87.38_{0.79}$ \\
Feature Propagation & $80.92_{1.29}$ & $\mathbf{83.49}_{1.54}$ & $85.33_{1.44}$ & $\mathbf{87.32}_{1.07}$ & $\mathbf{87.75}_{1.12}$ \\
\textbf{iN2V} (own) & $\mathbf{81.01}_{1.42}$ & $83.28_{1.41}$ & $85.92_{1.17}$ & $86.77_{1.42}$ & $87.60_{1.81}$ \\ 
Original features & $75.27_{2.63}$ & $83.37_{1.17}$ & $\mathbf{86.23}_{1.77}$ & $86.99_{0.86}$ & $87.05_{1.20}$ \\ \hline
\textcolor{gray}{N2V (transductive)} & \textcolor{gray}{$82.43_{0.77}$} & \textcolor{gray}{$84.45_{1.23}$} & \textcolor{gray}{$86.50_{1.04}$} & \textcolor{gray}{$86.85_{1.16}$} & \textcolor{gray}{$88.30_{1.47}$} \\
\bottomrule
\end{tabular}}
\label{tab:cat_sage_base}
\end{table}

\begin{table}[]
\centering
\caption{MLP on different iN2V setups concatenated with the original graph features.}
\resizebox{\linewidth}{!}{
\begin{tabular}{lrrrrr} \toprule
& \multicolumn{5}{l}{Percentage of training data} \\ \midrule
 Dataset & 10\% & 20\% & 40\% & 60\% & 80\% \\ \midrule
\multicolumn{6}{l}{Cora} \\ \hline
N2V (inductive) & $66.48_{1.68}$ & $70.54_{1.20}$ & $75.47_{1.91}$ & $76.09_{2.08}$ & $76.27_{1.68}$ \\
frozen ($\lambda = 1$) & $76.83_{1.66}$ & $79.70_{1.44}$ & $82.76_{1.87}$ & $85.31_{1.85}$ & $85.06_{2.20}$ \\
post-hoc & $\mathbf{82.01}_{1.38}$ & $82.53_{1.58}$ & $85.81_{1.38}$ & $85.63_{1.73}$ & $\mathbf{87.12}_{2.08}$ \\
p-h w losses & $81.85_{1.42}$ & $83.27_{1.21}$ & $\mathbf{85.83}_{1.45}$ & $\mathbf{86.27}_{1.97}$ & $86.68_{1.66}$ \\
p-h w sampling & $81.72_{1.26}$ & $\mathbf{83.97}_{1.35}$ & $85.37_{1.35}$ & $85.81_{1.77}$ & $86.49_{1.62}$ \\
\bottomrule
\end{tabular}}
\label{tab:cat_mlp_n2v}
\end{table}

\begin{table}[]
\centering
\caption{GraphSAGE on different iN2V setups concatenated with the original graph features.}
\resizebox{\linewidth}{!}{
\begin{tabular}{lrrrrr} \toprule
& \multicolumn{5}{l}{Percentage of training data} \\ \midrule
 Dataset & 10\% & 20\% & 40\% & 60\% & 80\% \\ \midrule
\multicolumn{6}{l}{Cora} \\ \hline
N2V (inductive) & $73.97_{3.35}$ & $81.92_{1.84}$ & $85.97_{1.67}$ & $\mathbf{86.97}_{0.96}$ & $87.38_{0.79}$ \\
frozen ($\lambda = 1$) & $76.56_{1.72}$ & $81.51_{1.77}$ & $85.12_{1.38}$ & $86.62_{1.87}$ & $87.60_{1.81}$ \\
post-hoc & $81.01_{1.42}$ & $81.98_{1.75}$ & $\mathbf{86.13}_{1.08}$ & $86.55_{1.35}$ & $86.64_{1.26}$ \\
p-h w losses & $\mathbf{81.45}_{1.39}$ & $82.82_{1.20}$ & $85.92_{1.17}$ & $86.35_{1.56}$ & $87.45_{1.39}$ \\
p-h w sampling & $80.97_{1.47}$ & $\mathbf{83.28}_{1.41}$ & $85.71_{1.44}$ & $86.77_{1.42}$ & $\mathbf{87.64}_{1.27}$ \\
\bottomrule
\end{tabular}}
\label{tab:cat_sage_n2v}
\end{table}


\begin{table}[]
\centering
\caption{Comparison of best iN2V variant vs baselines; GAT accuracy. Gray numbers are not directly comparable as they use additional information (graph features/transductive setup).}
\resizebox{\linewidth}{!}{
\begin{tabular}{lrrrrr} \toprule
& \multicolumn{5}{l}{Percentage of training data} \\ \midrule
 Dataset & 10\% & 20\% & 40\% & 60\% & 80\% \\ \midrule
\multicolumn{6}{l}{Cora} \\ \midrule
N2V (inductive) & $69.16_{5.02}$ & $77.14_{1.54}$ & $81.97_{1.67}$ & $83.47_{1.42}$ & $85.20_{1.53}$ \\
Feature Propagation & $79.10_{1.02}$ & $\mathbf{81.90}_{1.53}$ & $83.02_{1.75}$ & $84.30_{1.22}$ & $85.31_{2.43}$ \\
\textbf{iN2V} (own) & $\mathbf{79.70}_{1.29}$ & $81.74_{1.27}$ & $\mathbf{83.90}_{1.14}$ & $\mathbf{84.58}_{1.05}$ & $\mathbf{85.68}_{2.04}$ \\ \hline
\textcolor{gray}{Original features} & \textcolor{gray}{$81.35_{1.83}$} & \textcolor{gray}{$84.38_{1.13}$} & \textcolor{gray}{$86.51_{1.19}$} & \textcolor{gray}{$88.21_{0.84}$} & \textcolor{gray}{$88.15_{1.13}$} \\
\textcolor{gray}{N2V (transductive)} & \textcolor{gray}{$79.20_{1.53}$} & \textcolor{gray}{$81.89_{1.17}$} & \textcolor{gray}{$84.16_{1.38}$} & \textcolor{gray}{$84.98_{0.87}$} & \textcolor{gray}{$85.94_{1.67}$} \\
\bottomrule
\end{tabular}}
\label{tab:full_gat_base}
\end{table}

\begin{table}[]
\centering
\caption{Comparison of best iN2V variant vs baselines; GIN accuracy. Gray numbers are not directly comparable as they use additional information (graph features/transductive setup).}
\resizebox{\linewidth}{!}{
\begin{tabular}{lrrrrr} \toprule
& \multicolumn{5}{|c}{Percentage of training data} \\ \midrule
 Dataset & 10\% & 20\% & 40\% & 60\% & 80\% \\ \midrule
\multicolumn{6}{l}{Cora} \\ \midrule
N2V (inductive) & $68.38_{2.63}$ & $74.48_{2.55}$ & $81.76_{1.10}$ & $83.97_{1.39}$ & $\mathbf{85.20}_{1.92}$ \\
Feature Propagation & $78.27_{1.37}$ & $81.02_{1.56}$ & $82.40_{1.72}$ & $84.32_{1.10}$ & $85.13_{1.98}$ \\
\textbf{iN2V} (own) & $\mathbf{79.58}_{1.16}$ & $\mathbf{81.21}_{1.24}$ & $\mathbf{84.03}_{1.25}$ & $\mathbf{84.91}_{1.10}$ & $85.06_{1.67}$ \\ \hline
\textcolor{gray}{Original features} & \textcolor{gray}{$82.40_{1.08}$} & \textcolor{gray}{$84.05_{0.79}$} & \textcolor{gray}{$86.70_{1.39}$} & \textcolor{gray}{$88.08_{1.05}$} & \textcolor{gray}{$87.34_{0.95}$} \\
\textcolor{gray}{N2V (transductive)} & \textcolor{gray}{$79.09_{1.35}$} & \textcolor{gray}{$81.56_{1.35}$} & \textcolor{gray}{$83.23_{1.26}$} & \textcolor{gray}{$85.31_{0.85}$} & \textcolor{gray}{$86.09_{1.85}$} \\
\bottomrule
\end{tabular}}
\label{tab:full_gin_base}
\end{table}

\begin{table}[]
\centering
\caption{Comparison of N2V vs different iN2V setups, GAT accuracy.}
\resizebox{\linewidth}{!}{
\begin{tabular}{lrrrrr} \toprule
& \multicolumn{5}{l}{Percentage of training data} \\ \midrule
 Dataset & 10\% & 20\% & 40\% & 60\% & 80\% \\ \midrule
\multicolumn{6}{l}{Cora} \\ \midrule
N2V (inductive) & $69.16_{5.02}$ & $77.14_{1.54}$ & $81.97_{1.67}$ & $83.47_{1.42}$ & $85.20_{1.53}$ \\
frozen ($\lambda = 1$) & $75.96_{1.82}$ & $80.49_{1.61}$ & $82.91_{1.18}$ & $84.45_{1.08}$ & $85.06_{1.62}$ \\
post-hoc & $79.52_{1.25}$ & $81.15_{1.59}$ & $83.85_{0.99}$ & $84.21_{1.31}$ & $85.42_{1.91}$ \\
p-h w losses & $79.16_{1.16}$ & $81.26_{1.26}$ & $\mathbf{83.90}_{1.14}$ & $84.35_{1.54}$ & $\mathbf{85.68}_{2.04}$ \\
p-h w sampling & $\mathbf{79.70}_{1.29}$ & $\mathbf{81.74}_{1.27}$ & $83.44_{0.69}$ & $\mathbf{84.58}_{1.05}$ & $85.35_{1.65}$ \\
\bottomrule
    \end{tabular}}
    \label{tab:full_gat_setups}
\end{table}

\begin{table}[]
\centering
\caption{Comparison of N2V vs different iN2V setups, GIN accuracy.}
\resizebox{\linewidth}{!}{
\begin{tabular}{lrrrrr} \toprule
& \multicolumn{5}{l}{Percentage of training data} \\ \midrule
 Dataset & 10\% & 20\% & 40\% & 60\% & 80\% \\ \midrule
\multicolumn{6}{l}{Cora} \\ \midrule
N2V (inductive) & $68.38_{2.63}$ & $74.48_{2.55}$ & $81.76_{1.10}$ & $83.97_{1.39}$ & $85.20_{1.92}$ \\
frozen ($\lambda = 1$) & $75.57_{2.08}$ & $79.45_{1.36}$ & $82.23_{1.26}$ & $83.69_{1.19}$ & $84.58_{2.36}$ \\
post-hoc & $79.42_{1.37}$ & $80.66_{1.64}$ & $\mathbf{84.03}_{1.25}$ & $84.10_{1.52}$ & $85.09_{2.21}$ \\
p-h w losses & $79.51_{1.10}$ & $\mathbf{81.21}_{1.46}$ & $83.53_{1.30}$ & $83.71_{1.70}$ & $\mathbf{85.72}_{1.86}$ \\
p-h w sampling & $\mathbf{79.58}_{1.16}$ & $81.21_{1.24}$ & $83.15_{0.74}$ & $\mathbf{84.91}_{1.10}$ & $85.06_{1.67}$ \\
\end{tabular}}
    \label{tab:full_gin_setups}
\end{table}

\begin{table}[]
\centering
\caption{Comparison of best iN2V variant vs baselines concatenated with the original graph features; GAT accuracy. Gray numbers are not directly comparable as they use additional information (transductive setup).}
\resizebox{\linewidth}{!}{
\begin{tabular}{lrrrrr} \toprule
& \multicolumn{5}{l}{Percentage of training data} \\ \midrule
 Dataset & 10\% & 20\% & 40\% & 60\% & 80\% \\ \midrule
\multicolumn{6}{l}{Cora} \\ \midrule
N2V (inductive) & $80.65_{2.14}$ & $83.12_{1.22}$ & $86.08_{1.52}$ & $87.25_{1.09}$ & $\mathbf{88.27}_{1.48}$ \\
Feature Propagation & $81.56_{1.14}$ & $84.08_{1.37}$ & $86.49_{1.33}$ & $\mathbf{88.12}_{1.36}$ & $87.56_{1.35}$ \\
\textbf{iN2V} (own) & $\mathbf{81.68}_{1.23}$ & $\mathbf{84.11}_{1.12}$ & $\mathbf{86.70}_{1.36}$ & $87.99_{0.85}$ & $87.93_{1.53}$ \\ 
Original features & $81.35_{1.83}$ & $84.38_{1.13}$ & $86.51_{1.19}$ & $88.21_{0.84}$ & $88.15_{1.13}$ \\ \hline
\textcolor{gray}{N2V (transductive)} & \textcolor{gray}{$82.69_{1.21}$} & \textcolor{gray}{$84.90_{1.08}$} & \textcolor{gray}{$86.93_{1.38}$} & \textcolor{gray}{$87.68_{1.09}$} & \textcolor{gray}{$88.27_{1.29}$} \\
\bottomrule
\end{tabular}}
\label{tab:cat_gat_base}
\end{table}

\begin{table}[]
\centering
\caption{Comparison of best iN2V variant vs baselines concatenated with the original graph features; GIN accuracy. Gray numbers are not directly comparable as they use additional information (transductive setup).}
\resizebox{\linewidth}{!}{
\begin{tabular}{lrrrrr} \toprule
& \multicolumn{5}{l}{Percentage of training data} \\ \midrule
 Dataset & 10\% & 20\% & 40\% & 60\% & 80\% \\ \midrule
\multicolumn{6}{l}{Cora} \\ \midrule
N2V (inductive) & $81.53_{1.44}$ & $\mathbf{84.26}_{1.18}$ & $\mathbf{86.53}_{1.10}$ & $87.40_{2.26}$ & $87.38_{1.40}$ \\
Feature Propagation & $\mathbf{82.54}_{1.40}$ & $84.16_{1.34}$ & $85.83_{1.44}$ & $\mathbf{87.44}_{1.34}$ & $87.38_{1.42}$ \\
\textbf{iN2V} (own) & $81.68_{1.23}$ & $83.80_{1.02}$ & $86.51_{1.12}$ & $87.32_{1.01}$ & $\mathbf{87.79}_{1.43}$ \\ 
Original features & $82.40_{1.08}$ & $84.05_{0.79}$ & $86.70_{1.39}$ & $88.08_{1.05}$ & $87.34_{0.95}$ \\ \hline
\textcolor{gray}{N2V (transductive)} & \textcolor{gray}{$82.42_{1.11}$} & \textcolor{gray}{$84.28_{0.89}$} & \textcolor{gray}{$86.43_{1.21}$} & \textcolor{gray}{$87.21_{1.06}$} & \textcolor{gray}{$88.34_{1.33}$} \\
\bottomrule

\end{tabular}}
\label{tab:cat_gin_base}
\end{table}

\begin{table}[]
\centering
\caption{GAT on different iN2V setups concatenated with the original graph features.}
\resizebox{\linewidth}{!}{
\begin{tabular}{lrrrrr} \toprule
& \multicolumn{5}{l}{Percentage of training data} \\ \midrule
 Dataset & 10\% & 20\% & 40\% & 60\% & 80\% \\ \midrule
\multicolumn{6}{l}{Cora} \\ \midrule
N2V (inductive) & $80.65_{2.14}$ & $83.12_{1.22}$ & $86.08_{1.52}$ & $87.25_{1.09}$ & $\mathbf{88.27}_{1.48}$ \\
frozen ($\lambda = 1$) & $79.19_{1.56}$ & $82.24_{1.62}$ & $85.43_{1.39}$ & $86.99_{1.34}$ & $87.93_{1.53}$ \\
post-hoc & $81.75_{1.27}$ & $83.06_{1.67}$ & $\mathbf{86.70}_{1.36}$ & $86.85_{1.46}$ & $87.75_{1.87}$ \\
p-h w losses & $\mathbf{81.78}_{1.67}$ & $83.55_{1.46}$ & $86.53_{0.93}$ & $87.45_{1.18}$ & $87.12_{1.99}$ \\
p-h w sampling & $81.68_{1.23}$ & $\mathbf{84.11}_{1.12}$ & $86.16_{1.14}$ & $\mathbf{87.99}_{0.85}$ & $88.08_{1.58}$ \\
\bottomrule
\end{tabular}}
\label{tab:cat_gat_n2v}
\end{table}

\begin{table}[]
\centering
\caption{GIN on different iN2V setups concatenated with the original graph features.}
\resizebox{\linewidth}{!}{
\begin{tabular}{lrrrrr} \toprule
& \multicolumn{5}{l}{Percentage of training data} \\ \midrule
 Dataset & 10\% & 20\% & 40\% & 60\% & 80\% \\ \midrule
\multicolumn{6}{l}{Cora} \\ \midrule
N2V (inductive) & $81.53_{1.44}$ & $\mathbf{84.26}_{1.18}$ & $86.53_{1.10}$ & $\mathbf{87.40}_{2.26}$ & $87.38_{1.40}$ \\
frozen ($\lambda = 1$) & $79.73_{1.69}$ & $82.98_{1.49}$ & $85.44_{1.28}$ & $86.42_{1.78}$ & $\mathbf{87.79}_{1.43}$ \\
post-hoc & $\mathbf{81.68}_{1.23}$ & $82.83_{1.68}$ & $86.51_{1.12}$ & $86.92_{1.42}$ & $87.49_{1.74}$ \\
p-h w losses & $81.58_{1.34}$ & $83.34_{1.56}$ & $\mathbf{86.64}_{1.07}$ & $86.62_{1.57}$ & $87.60_{1.59}$ \\
p-h w sampling & $81.66_{1.41}$ & $83.80_{1.02}$ & $85.95_{0.74}$ & $87.32_{1.01}$ & $87.79_{1.60}$ \\
\bottomrule
\end{tabular}}
\label{tab:cat_gin_n2v}
\end{table}

\end{document}